\PassOptionsToPackage{table}{xcolor}
\documentclass[11pt]{article}

\usepackage{acl}

\usepackage{enumitem} 
\usepackage{times}
\usepackage{latexsym}
\usepackage{booktabs}
\usepackage[linesnumbered,ruled,vlined]{algorithm2e}
\usepackage{xcolor}
\definecolor{halfgray}{gray}{0.55}
\usepackage{complexity}
\usepackage[T1]{fontenc}
\usepackage{booktabs}
\usepackage{tabularx}
\usepackage[utf8]{inputenc}
\usepackage{enumitem}  
\usepackage{microtype}

\usepackage{inconsolata}

\usepackage{adjustbox}
\usepackage{amsmath,amssymb}
\usepackage{natbib}
\usepackage{graphicx}
\usepackage{tikz}
\usetikzlibrary{arrows.meta,positioning,fit,calc}
\usepackage{svg}
\usepackage{booktabs}
\usepackage{multirow,makecell}
\usepackage{caption}
\usepackage{tabularx}
\usepackage{CJKutf8}
\usepackage{hyperref}
\usepackage{cleveref}
\usepackage[most]{tcolorbox}

\tcbset{colback=white}
\crefname{tcb@cnt@problemwithtitle}{Problem}{Problems}
\tcbuselibrary{listings, skins, breakable}
\definecolor{pblue}{rgb}{0.13,0.13,1}
\definecolor{pgreen}{rgb}{0,0.5,0}
\lstdefinelanguage{cofola}{
    language=Python,
    basicstyle=\ttfamily\footnotesize,
    keywordstyle=\bfseries\color{pblue},
    stringstyle=\bfseries\itshape\color{green!40!black},
    commentstyle=\itshape\color{black!60},
    showspaces=false,
    numbers=left,
    numberstyle=\tiny\color{halfgray},
    breaklines=true,
    showstringspaces=false,
    tabsize=1,
    emph={
        set, bag, choose, choose_replace, in,
        func, disjoint, subset, seq, cir,
        permute, choose_permute, choose_replace_permute, circular_permute, par, comp, partition, ordered_partition, tup
    },
    emphstyle={\bfseries\color{pblue}},
}
\lstdefinestyle{verbatim}{
    basicstyle=\ttfamily, 
    literate=
    *{\\}{{\textbackslash{}}}{1}, 
    keepspaces,
}
\newtcblisting[auto counter]{cofolacode}[1]{%
    label=#1,
    enhanced,
    frame hidden, 
    breakable,
    borderline north = {0.5pt}{0pt}{black},
    borderline south = {0.5pt}{0pt}{black},
    opacityfill=0,
    top=-5pt,bottom=-5pt,
    fonttitle=\bfseries,
    listing only,
    listing options={language=cofola}
}
\newtcblisting[auto counter]{cofolacodeblock}[1]{%
    label=#1,
    enhanced,
    frame hidden, 
    borderline north = {0.5pt}{0pt}{black},
    borderline south = {0.5pt}{0pt}{black},
    borderline west = {0.5pt}{0pt}{black},
    borderline east = {0.5pt}{0pt}{black},
    opacityfill=0,
    top=-5pt,bottom=-5pt,
    fonttitle=\bfseries,
    listing only,
    listing options={language=cofola}
}
\newtcblisting[auto counter]{problemwithcode}[2]{%
    label=#1,
    enhanced,
    colback=white,
    breakable,
    bottom=-5pt,left=0pt,right=0pt, middle=1pt,
    title={Code Example \thetcbcounter},
    fonttitle=\bfseries,
    comment and listing,
    comment={\textbf{Problem:} #2},
    after={\par\noindent},
    listing options={language=cofola}
}
\Crefname{tcb@cnt@errexample}{Error}{Errors}

\newtcbtheorem[auto counter]{errexample}{Error}%
{
    width=\textwidth,
    colback=gray!5,
    colframe=black!70,
    boxrule=0.8pt,
    arc=2mm,
    fonttitle=\bfseries\sffamily,
    fontupper=\ttfamily\small,
    left=6pt, right=6pt, top=6pt, bottom=6pt,
}{err}

\newtheorem{example}{Example}

\title{CombEval: A Framework for Evaluating Combinatorial Counting in Large Language Models}

\author{
  \textbf{Yuxu Zhou\textsuperscript{1}},
  \textbf{Ond\v{r}ej Ku\v{z}elka\textsuperscript{2}},
  \textbf{Yuyi Wang\textsuperscript{3,4}},
  \textbf{Yuanhong Wang\textsuperscript{1}\thanks{Corresponding authors.}},
  \textbf{Yi Chang\textsuperscript{1,5,6}\footnotemark[1]}
\\
  \textsuperscript{1}School of Artificial Intelligence, Jilin University, Changchun, China
\\
  \textsuperscript{2}Czech Technical University in Prague, Prague, Czech Republic
\\
  \textsuperscript{3}CRRC Zhuzhou Institute, Zhuzhou, China \quad
  \textsuperscript{4}Tengen Intelligence Institute, China
\\
  \textsuperscript{5}International Center of Future Science, Jilin University
\\
  \textsuperscript{6}Engineering Research Center of Knowledge-Driven Human-Machine Intelligence, MOE, China
\\
  \small{
    \textbf{Correspondence:} \href{mailto:lucienwang@jlu.edu.cn}{lucienwang@jlu.edu.cn}
  }
}

\begin{document}

\maketitle
\begin{abstract}
  We present CombEval, a dynamic benchmark for evaluating combinatorial counting in large language models. CombEval represents each problem as a typed Cofola specification over entities, combinatorial objects, object dependencies, and constraints, enabling controlled generation of natural-language counting problems with exact solver-verified answers. Unlike static collections, CombEval supports systematic variation of object type, entity scale, constraint count, and reasoning depth. We evaluate 11 LLMs under direct and code-augmented settings and find that models remain brittle on ordered objects, indistinguishable elements, relatively positional constraints, and nested object dependencies.  Error analysis further identifies failures in constraint interpretation and counting principles. CombEval provides a diagnostic testbed for studying when and why LLMs fail at combinatorial reasoning.
  The code and generated benchmark suites are publicly available at \url{https://github.com/YuxuZhou-CN/combination-problem-generation}.
\end{abstract}

\section{Introduction}

\emph{Combinatorial counting} (\emph{CO}) or \emph{enumeration} is a fundamental branch of mathematics \cite{stanley2011enumerative}, with one of its most critical functions lies in calculating probability. 
For most equiprobable events, probability is derived by using CO to count the number of favorable outcomes and the total number of possible outcomes, then computing the ratio of these two values. 
CO is also crucial in key domains such as finance, logistics, and healthcare, where it empowers constraint-driven decision optimization by quantifying viable options amid complex requirements. Yet its complexity stems primarily from the intricate nature of these constraints, which are often interdependent, dynamic, and multi-faceted, making it challenging to develop one-size-fits-all systematic solutions that adapt to diverse real-world scenarios. 
This is where large language models (LLMs) emerge as a promising approach to CO problems.
The strength of LLMs in abstract reasoning and processing unstructured, complex constraints allows them to tackle CO problems without relying on rigid, scenario-specific algorithms.

LLMs have demonstrated remarkable capabilities in tasks such as natural language processing \cite{zhao2023survey,kalyan2024survey}, 
code generation \cite{jiang2024survey}, and question answering \cite{kuang2025natural}, with their superior performance on mathematical reasoning benchmarks being particularly noteworthy \cite{xu2025towards}.
Current mainstream models~\cite{openai2025gptoss120bgptoss20bmodel,Qwen3,deepseekai2025deepseekv32} have achieved or even surpassed human-average accuracy on widely used mathematical reasoning datasets like GSM8K \cite{cobbe2021training} and MATH \cite{hendrycks2021measuring},
sparking widespread interest in exploring the potential of LLMs to tackle more complex mathematical problems and reasoning tasks \cite{yang2025position}.

Current benchmarks for evaluating the CO capabilities of LLMs are often contained within broader mathematical reasoning datasets \cite{hendrycks2021measuring,zheng2021minif2f,aime_1983_2024,balloccu2024benchmark,xuejun2025mathesis}.
For example, the MATH dataset \cite{hendrycks2021measuring} includes $1245$ problems in the "Counting and Probability" category, covering topics such as basic counting principles, permutations and combinations, and probability calculations.
There are also limited benchmarks specifically targeting CO problems, such as CombiBench \cite{liu2025combibench}, where $100$ CO problems are collected spanning from middle school to International Mathematical Olympiad levels.
Though facilitating initial assessment of LLMs' CO capabilities, these static benchmarks usually face the \emph{data contamination} and \emph{spurious reasoning} issues prevalent in mathematics benchmarks.

The data contamination issue arises when test questions or their variants are present in the training data of LLMs \cite{balloccu2024leak, golchin2023time, zhou2025lessleak}. 
This issue makes models potentially able to achieve high scores through memorization rather than genuine reasoning, leading to overestimation of their true reasoning capabilities \cite{deng-etal-2024-investigating}.
On the other hand, even when the benchmark problems are not directly present in the training data, models may still exploit superficial textual patterns to arrive at correct answers without engaging in deep logical reasoning \cite{mirzadeh2024gsm, lai2025making, boye2025large}.
In scenarios requiring structural abstraction like combinatorics, models might memorize the superficial association that "choosing $k$ items from $n$" corresponds to binomial coefficients, but their reasoning ability collapses rapidly when additional constraints are applied \cite{shrestha2025mathematical}. 

\begin{figure*}[tp]
  \centering
  \includegraphics[width=\linewidth]{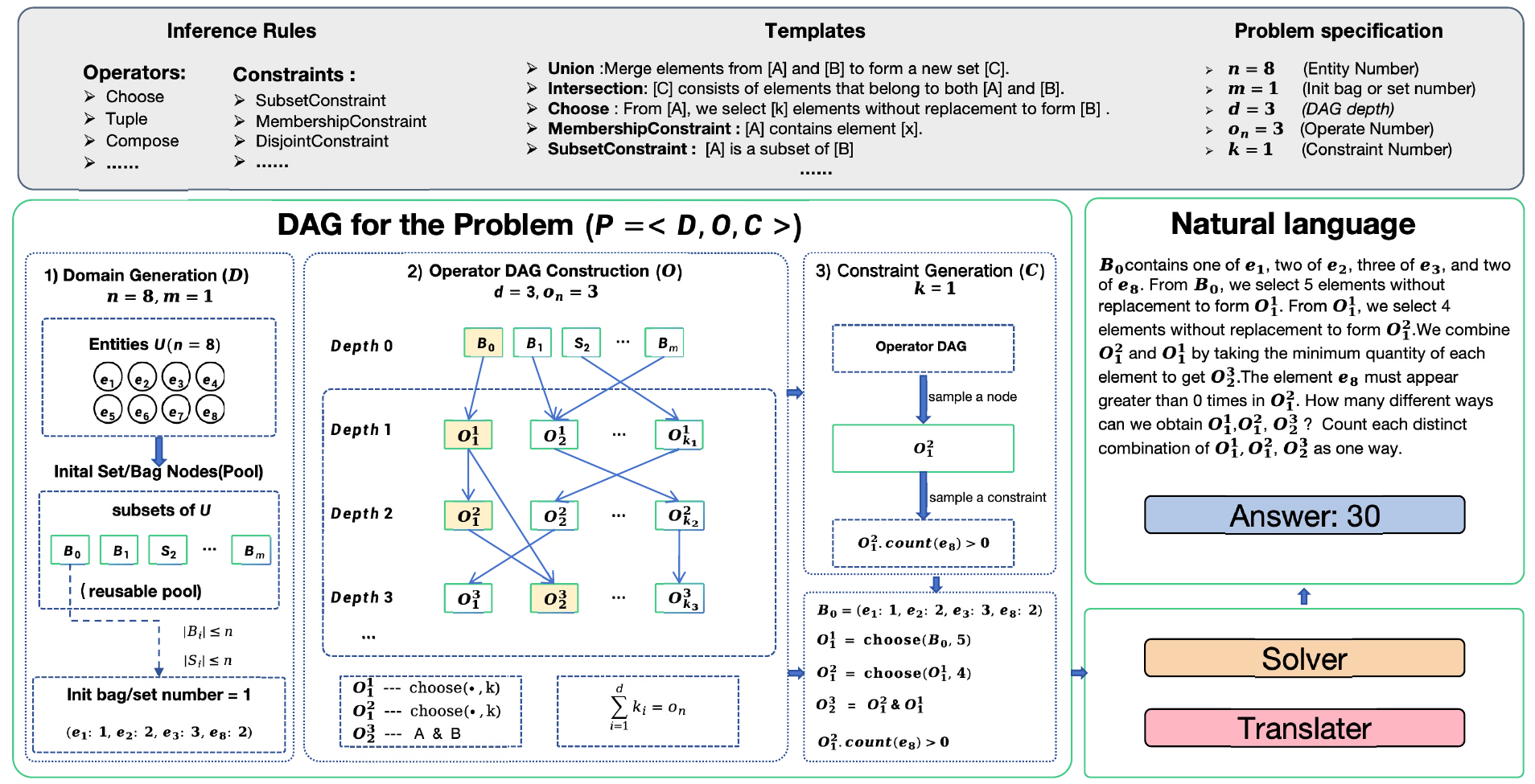}
  \caption{Overview of the CombEval generation framework with the current Cofola backend. The generator samples a typed object DAG $\mathcal{P}=\langle D,O,C\rangle$, attaches compatible constraints, verbalizes the resulting formal specification with templates, and verifies the exact answer using the Cofola solver.}
  \label{fig:methodarchitecture}
\end{figure*}

Therefore, this paper proposes CombEval, a dynamic evaluation framework for CO in LLMs.
CombEval synthesizes CO instances from a typed formal specification rather than from a fixed question bank, as outlined in \Cref{fig:methodarchitecture}.
This construction makes it possible to sample fresh problems after benchmark design time, while also controlling the structural variables that affect difficulty.
Moreover, because every instance is grounded in a formal program and solved exactly before evaluation, the benchmark can test whether models reason about the underlying structure rather than merely matching familiar surface templates.

At the core of CombEval is Cofola~\citep{wang2026solvingcombinatorialcountingproblems},
a typed declarative language and solver for combinatorial counting.
Following Cofola, we represent a problem as $\langle D, O, C\rangle$, where $D$ is a finite entity domain, $O$ is a set of combinatorial objects defined over $D$ (e.g., sets, sequences, partitions) with potential dependencies between them, and $C$ is a set of constraints over these objects.
Cofola provides seven object types including sets, bags, tuples, sequences, circles, partitions, and compositions, together with constraints for membership, subset and disjointness, multiplicity and cardinality, absolute tuple positions, relative order patterns, and group-level conditions (see \Cref{tab:cofola-operations,tab:constraints}).
Its solver compiles the formal specification into weighted first-order model counting (WFOMC) instances, which can be solved efficiently with off-the-shelf WFOMC solvers \cite{DBLP:conf/uai/BremenK21,kuzelka_weighted_2021,DBLP:journals/corr/abs-2507-19182}, enabling exact answer verification for a wide range of combinatorial problems.


Using CombEval, we construct several solver-verified evaluation suites, and conduct a systematic evaluation of $11$ mainstream general-purpose, math-specialized, and thinking-specialized LLMs, including both open-source models (e.g., Qwen, DeepSeek, LLaMA, and gpt-oss) and closed-source models (e.g., GPT-5 and Gemini-3).
The main observations are:

\begin{itemize}
\item Model scale and advanced reasoning techniques lead to improved overall accuracy under both zero-shot and code-augmented settings. However, all evaluated models still show clear degradation on typed objects that require indistinguishable elements, ordered structures, or multi-step dependencies.

\item CombEval enables fine-grained difficulty control. Increasing entity size, constraint count, or object-dependency depth predictably lowers model accuracy, showing that the generated suites support granular model comparisons rather than only a single aggregate score.

\item Surface template variations can cause performance fluctuations to some extent, especially for less capable (but still strong) models.
However, the latest \textbf{gpt-5.5}, one of the strongest models, show greater robustness to these variations, suggesting that their reasoning capabilities are not tightly coupled to specific prompt formulations.
\end{itemize}

\section{Related Work}

Mathematical reasoning benchmarks such as GSM8K \cite{cobbe2021training} and MATH \cite{hendrycks2021measuring} have established a foundation for evaluating arithmetic and algebraic capabilities including combinatorial counting in LLMs, but they also face significant limitations. Primarily, these static datasets are highly susceptible to data contamination.
As pre-training corpora expand, test questions are increasingly memorized rather than solved, distorting evaluation results \cite{balloccu2024leak, golchin2023time, zhou2025lessleak}. 
While researchers have attempted to mitigate this through dynamic generation in other reasoning fields \cite{shi2022stepgame,saparov2023testing,wan2024logicasker,opedal2024mathgap,wang2025generating,ariyani2025there}, constructing benchmarks specifically for CO presents unique challenges. 
First, CO problems possess high structural and semantic diversity, making it difficult to design a unified framework that covers the full spectrum of problem types. 
Second, efficient solutions often require specialized mathematical tools, such as generating functions and inclusion-exclusion principles \cite{stanley2011enumerative,pak_complexity_2019,ferraris2015counting}, which complicates the design of automated answer verification. 
Finally, as many CO problems go beyond polynomial-time complexity \cite{valiant_complexity_1979}, generation and evaluation frameworks must be carefully designed to ensure scalability without resorting to brute-force enumeration.

Solver-backed benchmark generation therefore depends not only on natural-language templates, but also on the expressiveness and reliability of the formal backend.
There are various existing methods for automatically solving CO problems, ranging from Constraint Satisfaction Programming (CSP) \cite{akgun2022conjure}, Answer Set Programming (ASP) \cite{gebser2022answer}, to lifted inference techniques \cite{van_den_broeck_lifted_2011,taghipour2012lifted,kuzelka_weighted_2021}.
We provide a brief overview of these methods in \Cref{app:Existing_Solvers_for_Combinatorial_Counting}.
Recently, \citet{totis_lifted_2023} proposed CoLa/CoSo, a lifted framework for modeling and solving CO problems.
CoLa/CoSo provides an important lifted approach to combinatorial counting, but it is centered on a more restricted single-configuration language, where only one combinatorial object can be defined and solved per problem.
The updated Cofola language and solver~\citep{wang2026solvingcombinatorialcountingproblems} used in CombEval extends this line with typed object dependencies, sets and bags, ordered objects, grouped objects, and WFOMC-based solving.
This broader formal layer allows CombEval to generate larger and more structurally varied suites while retaining exact answer verification.

\section{Formalization of Combinatorial Counting Problems}
CombEval follows the updated Cofola semantics and formalizes each CO problem as a triple
$\mathcal{P}=\langle D,O,C\rangle$.
Here $D$ is a finite domain of entities, $O$ is a set of typed combinatorial objects over $D$, and $C$ is a set of constraints over these objects.
The answer is the number of valid combinatorial structures, i.e., assignments of concrete instances to every object in $O$ that respect the object dependencies and satisfy all constraints in $C$.
We introduce the components below with the following running example.

\begin{example}
\label{ex:running}
I have seven books on a shelf. 
Two are math books and one is a physics book. 
How many ways can I select $5$ books and arrange them if 1) the two math books and one physics book must be included, 2) the math books must be adjacent, and 3) both math books must be to the left of the physics book?
\end{example}

\paragraph{Entity domain}
The entity domain $D$ contains the atomic objects that may appear in the count.
In Example~\ref{ex:running}, $D$ contains seven book entities, including two math books $M_1, M_2$, one physics book $P$, and four other books $O_1, \dots, O_4$.
Entities may be distinguishable, as in ordinary sets, or may appear with multiplicity inside bags (multisets), where indistinguishable copies must not be over-counted.

\paragraph{Combinatorial objects}
The object set $O$ contains the combinatorial structures defined over $D$ that are relevant to the problem.
Each object is defined either directly from $D$ or derived from other objects through combinatorial operations.
For instance, for the problem in Example~\ref{ex:running}, $O$ contains a set object defined as $S = \texttt{set}(M_1, M_2, P, O_1, \dots, O_4)$, a subset object defined as $S' = \texttt{choose}(S, 5)$, and a sequence object defined as $T = \texttt{sequence}(S')$.
\Cref{tab:cofola-operations} summarizes the object types and representative operators available in Cofola.











\begin{table*}[tbp]
\centering
\caption{Cofola object types and representative operators used by CombEval. Sets and bags are basic objects that can be defined directly from the entity domain, while tuples, sequences, circles, partitions, and compositions are derived from existing set or bag objects.}
\label{tab:cofola-operations}
\begin{adjustbox}{max width=\textwidth}
\begin{tabular}{lll}
\toprule
\textbf{Object Type} & \textbf{Representative Operators} & \textbf{Description} \\
\midrule
Set & \texttt{set}, \texttt{choose}, \texttt{union}, \texttt{intersect}, \texttt{diff}, \texttt{supp} & Unordered entities and set operations. \\
Bag & \texttt{bag}, \texttt{choose}, \texttt{choose\_replace}, \texttt{add\_union}, \texttt{union}, \texttt{intersect}, \texttt{diff} & Multisets and replacement-based choices. \\
Tuple & \texttt{tuple}, \texttt{choose\_tuple}, \texttt{choose\_replace\_tuple} & Ordered arrangements supporting absolute-position constraints. \\
Sequence & \texttt{sequence}, \texttt{choose\_sequence}, \texttt{choose\_replace\_sequence} & Linear arrangements supporting relative-position patterns. \\
Circle & \texttt{circle}, \texttt{choose\_circle}, \texttt{choose\_replace\_circle} & Circular arrangements, optionally identifying reflection. \\
Partition & \texttt{partition} & Unordered grouping into disjoint parts. \\
Composition & \texttt{compose}, \texttt{index} & Ordered grouping into disjoint parts and access to a specific group. \\
\bottomrule
\end{tabular}
\end{adjustbox}
\end{table*}

\paragraph{Constraints}
The constraint set $C$ restricts the valid configurations of objects in $O$.
Cofola supports constraints over basic, ordered, and grouped objects, as summarized in \Cref{tab:constraints}.
For Example~\ref{ex:running}, the constraints include \texttt{subset} constraints to ensure the required books are included, a \texttt{together} constraint to enforce adjacency, and a \texttt{precedence} constraint to ensure the math books are to the left of the physics book.
The language also allows logical combinations of atomic constraints, enabling negated or disjunctive conditions when needed.

\begin{table*}[htbp]
\centering
\caption{Representative constraint families supported by the updated Cofola backend.}
\label{tab:constraints}
\begin{adjustbox}{max width=\textwidth}
\begin{tabular}{lll}
\toprule
\textbf{Object Scope} & \textbf{Constraint Family} & \textbf{Description} \\
\midrule
Set/Bag & Membership, subset, equality, disjointness & Require entities or objects to be included, excluded, equal, nested, or disjoint. \\
Set/Bag & Cardinality and multiplicity & Constraints over cardinalities and multiset multiplicities. \\
Tuple & Absolute-position constraints & Constrain a fixed tuple position, e.g., \texttt{T[i] == e} or \texttt{T[i] in S}. \\
Tuple & Count and deduplicated-count constraints & Count positions occupied by elements, or count distinct appearing elements. \\
Sequence/Circle & Relative-position patterns & Express together, precedence, immediate predecessor, and adjacency patterns. \\
Sequence/Circle & Pattern-count constraints & Count occurrences of relative-position patterns. \\
Partition/Composition & Group-level and indexed constraints & Constrain every group, or constrain a specific indexed group in a composition. \\
\bottomrule
\end{tabular}
\end{adjustbox}
\end{table*}

Putting these components together, Example~\ref{ex:running} can be expressed as the following Cofola program:
\begin{cofolacode}{}
math_books = set(math1, math2)
physics_book = set(physics)
books = set(math1, math2, physics, book4, book5, book6, book7)
choice = choose(books, 5)
math_books subset choice
physics_book subset choice
shelf = sequence(choice)
together(math_books) in shelf
math_books < physics_book in shelf
\end{cofolacode}

We do not expect that CombEval covers all CO problems.
More complex domains, such as graph objects, grid placements, arithmetic over entity-level numeric attributes, and arbitrary symmetry groups, would require additional language constructs and hand-crafted approaches; some of these extensions run into known \#\P-hard barriers \cite{valiant_complexity_1979}.

\section{Combinatorial Counting Instance Generation}

This section details the generation method for CO instances that we use in this work.

\paragraph{Problem Generation}

The generation of a CO problem follows from its formalization $\mathcal{P} = \langle D, O, C\rangle$.
As shown in Figure~\ref{fig:methodarchitecture}, our framework generates problems through a typed object-DAG pipeline.

First, we initialize a finite entity domain with $n$ entities and sample $m$ initial set or bag objects, forming the reusable source pool.
Next, we construct a Cofola object DAG by repeatedly sampling type-compatible operators, such as \texttt{choose}, \texttt{choose\_replace}, set/bag operations, \texttt{tuple}, \texttt{sequence}, \texttt{circle}, \texttt{compose}, \texttt{partition}, and indexing.
The maximum dependency depth $d$ and operator count $o_n$ control the length and breadth of the generated reasoning chain.
Finally, we sample $k$ constraints whose types match the selected objects, covering structural constraints, cardinality and multiplicity constraints, absolute-position constraints, relative-position patterns, and group-level constraints.


\paragraph{Natural Language Translation}

We use a template-based method to convert the generated CO problem into natural language.
Each Cofola object declaration and constraint is verbalized by a template associated with its type.
For example, a \texttt{choose} declaration becomes a sentence describing a selection without replacement, a \texttt{sequence} declaration becomes an arrangement in a line, and a \texttt{compose} declaration becomes a distribution into labeled groups.
The final query is generated from the target object or object set being counted, e.g., "How many different ways can we obtain the resulting objects?"
The full template library is provided in \Cref{app:Template_section}.

\paragraph{Problem Solving}

We use the Cofola solver as our backend because it supports a broader typed object language, multiple dependent objects, and a WFOMC-based compilation pipeline.
For each generated instance, Cofola computes the exact answer under a timeout budget; instances with solver failures, unsupported features, or unstable answers are filtered out before LLM evaluation.

\section{Experiments}

In this section, we present experiments evaluating the performance of various LLMs on the CombEval dataset generated using the proposed framework. 
Our experimental evaluation is systematically designed to address the following three primary research questions:
\begin{itemize}
      \item \textbf{RQ1: Capabilities of LLMs in Combinatorial Reasoning.} Can the generated combinatorial counting problems in CombEval faithfully reflect the capability hierarchy of current LLMs, producing performance trends consistent with existing benchmarks where stronger models achieve higher accuracy?
    \item \textbf{RQ2: Controllability of Problem Difficulty.} Can our generation framework effectively and predictably regulate the reasoning difficulty of problem instances by tuning multiple structural parameters?
    \item \textbf{RQ3: Impact of Prompt Templates on LLM Reasoning.} Does the surface form of custom problem templates bias or constrain the models' mathematical reasoning performance when underlying logical structures are kept strictly identical?
\end{itemize}
\subsection{Experimental Setup}

\paragraph{Models. } 

We evaluate a set of LLMs on the CombEval dataset, including both open-source and closed-source systems. 
The open-source models comprise Meta-Llama-3-8B-Instruct, DeepSeek-Math-7B-Instruct \cite{DeepSeek-Math-7B-Instruction}, DeepSeek-V3.2 \cite{deepseekai2025deepseekv32},
Qwen2.5-Math-7B-Instruct \cite{Qwen2.5-Math-7B-Instruct}, 
Qwen3-4B-Thinking-2507, 
Qwen3-30B-Thinking-2507 \cite{Qwen3}, 
gpt-oss-20B \cite{openai2025gptoss120bgptoss20bmodel}, and LLaMA-3.3-70B-Instruct \cite{dubey2024llama}, 
covering math-specialized, instruction-tuned, and reasoning-enhanced architectures across different scales. 
For closed-source models, we evaluate the state-of-the-art models gpt-5.1, gpt-5.5 and gemini-3-flash-preview-thinking.
Closed-source models are evaluated through their official APIs, while open-source models are run locally on a machine with 4 NVIDIA A40 GPUs.
All models are used under their licensed terms for research purposes.

\paragraph{Evaluation Protocol. }
We use a zero-shot setting with only the problem description and the required output format. Temperature is set to 0 for deterministic outputs, with a maximum generation length of 4096 tokens. 
The full prompt templates used for evaluation are provided in Appendix~\ref{app:Evaluation_Prompts}.

\paragraph{Metrics.} Accuracy is the primary metric. We use regular expressions to extract the final numerical answer; failure or mismatch is counted as an error.

For ease of analysis, we categorize the Cofola objects into four main types: \textsc{Choose} (objects derived from \texttt{choose} or \texttt{choose\_replace}), \textsc{Group} (objects derived from \texttt{compose} or \texttt{partition}), \textsc{Tuple} (objects derived from \texttt{tuple} or \texttt{choose\_tuple}), and \textsc{Sequence} (objects derived from \texttt{sequence} or \texttt{choose\_sequence}).

\subsection{RQ1: Capabilities of LLMs in Combinatorial Reasoning}
Using CombEval, we vary the key difficulty parameters: constraint count $k$ from 0 to 3, entity size $n$ over \{6,8,10,12\}, and number of combinatorial operations $o_n$ from 1 to 4. We fix $m=1$ and leave the DAG depth $d$ unconstrained, allowing the number of operations per layer to vary freely. With a 2000-second Cofola timeout to filter unstable instances, we construct 1,500 high-quality CombEval problems, whose type distribution is reported in the appendix~\ref{app:datasetdistribution}.
For a sanity check, we also use CoLa/CoSo to solve the same problem set, where $705$ problems are solvable by CoLa/CoSo that produce the same answers as Cofola.


To comprehensively assess reasoning capabilities, we evaluate the models under two distinct paradigms: standard natural language reasoning (Zero-shot) and code-augmented reasoning, where models are prompted to return executable Python code to compute the final answer. 
The Python setting is designed to test whether models can leverage programmatic reasoning to handle complex combinatorial structures.

\begin{figure}[t]
    \centering
    \includegraphics[width=\linewidth]{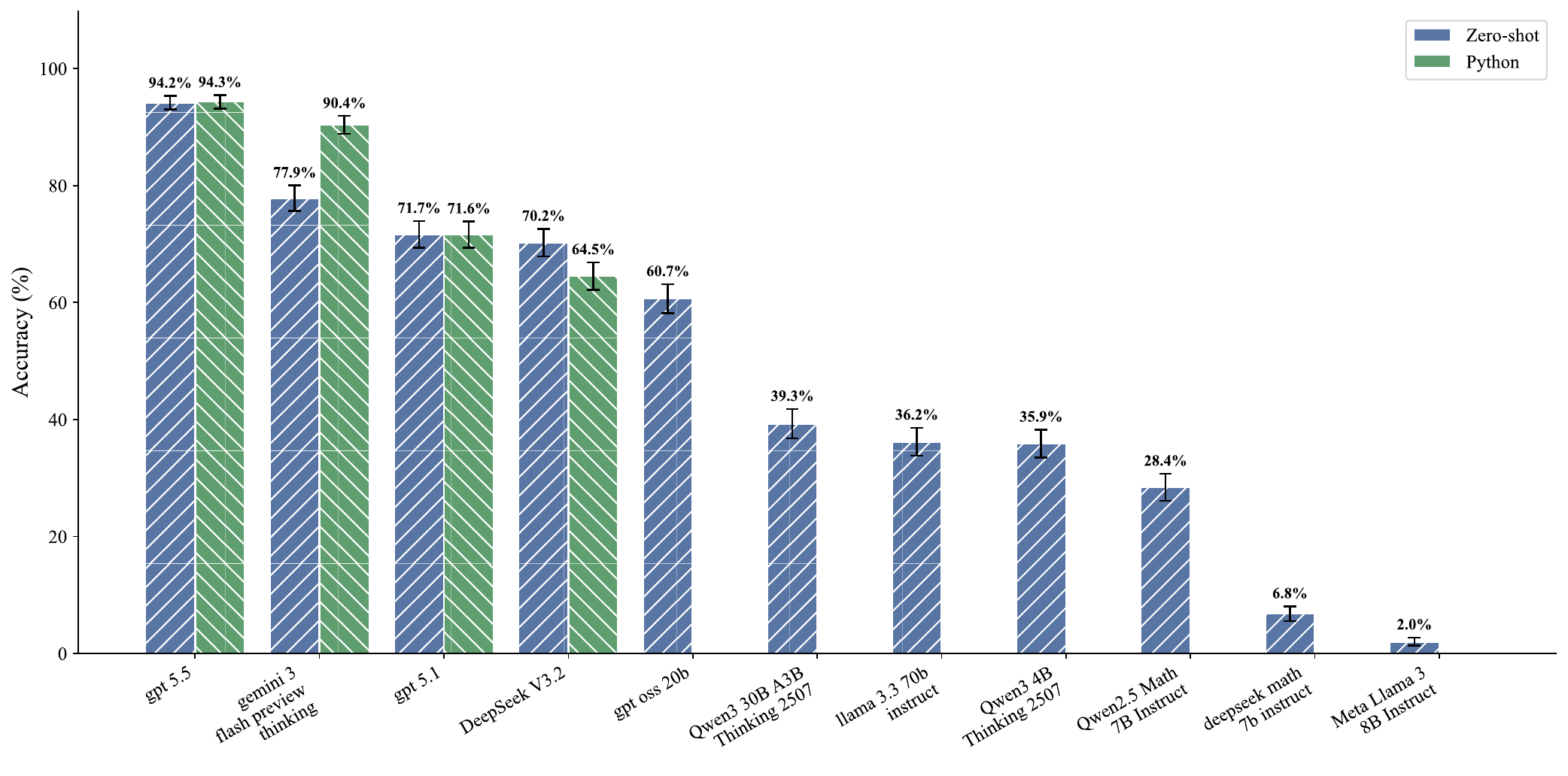}
    \caption{Overall accuracy comparison of evaluated LLMs on CombEval. Models were tested using standard natural language reasoning (Zero-shot) and code-augmented reasoning (Python). Due to the exceedingly weak coding capabilities of smaller models, which frequently fail to return compliant and executable code, they were not evaluated under the Python setting.}
    \label{fig:overall_accuracy_bar}
\end{figure}

As shown in Figure~\ref{fig:overall_accuracy_bar}, the generated evaluation problems exhibit strong discriminative power and clearly characterize the capability hierarchy of current LLMs. Stronger models with larger parameter scales or advanced reasoning mechanisms, such as \textbf{gpt-5.5} and  \textbf{gemini-3-flash-preview-thinking}, achieve substantial performance advantages in both settings, whereas smaller models show clear limitations. This performance ladder not only validates CombEval as an effective benchmark, but also directly reveals the capability boundaries of different LLM tiers when facing structured mathematical reasoning.

\subsection{RQ2: Controllability of Problem Difficulty}

To verify whether the generation parameters selected in our framework can effectively and predictably regulate problem difficulty, we adopt a controlled-variable design and analyze the impact of entity size, constraint count, and reasoning depth on LLM performance.
We also conduct experiments on the runtime of the backend CO solver under different $d, n, k$.
The results, shown in \Cref{app:Runtime_as_Difficulty_Indicator}, indicate that $n$ and $k$ have a significant impact on the solving time, while $d$ has a moderate effect.
Though, the impact of $d$ may be more pronounced when considering the basic combinatorial reasoning ability of LLMs, deeper depth often requires more complex multi-step reasoning.

\subsubsection{Impact of Operation Type and Difficulty Parameters}

\begin{figure*}[t]
    \centering
    \includegraphics[width=\linewidth]{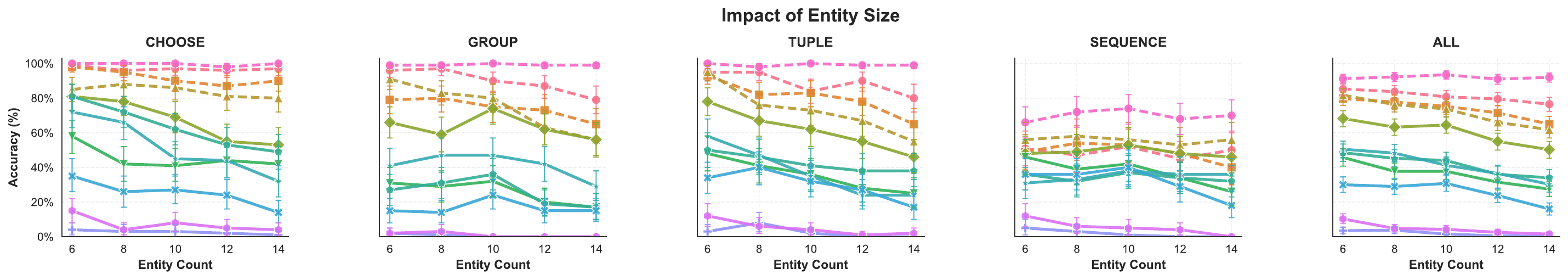}
    \includegraphics[width=\linewidth]{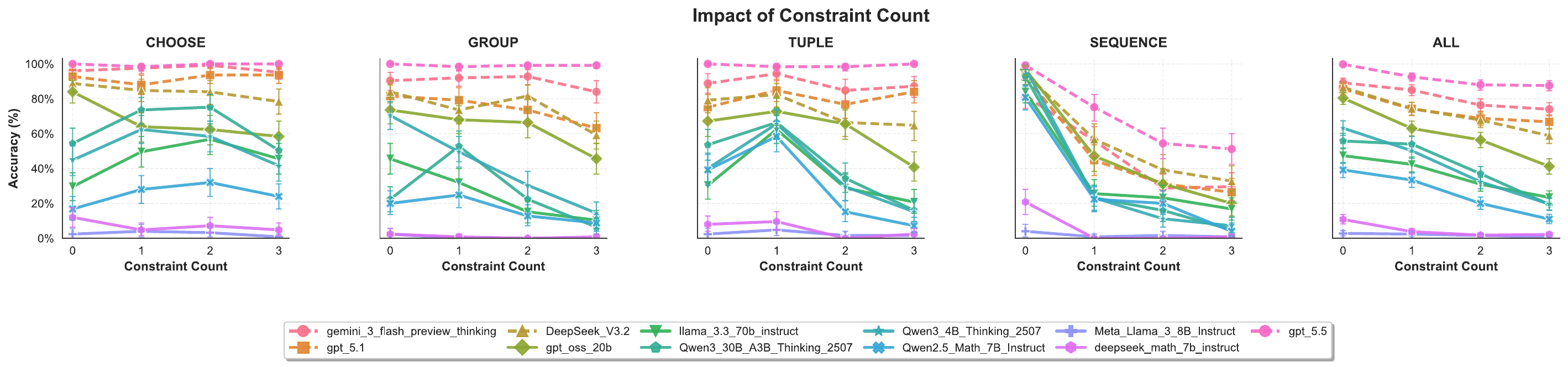}
    \caption{Impact of entity size and constraint count on model performance.}
    \label{fig:entity_constraint_analysis}
\end{figure*}

We analyze operation type, constraint count, and entity scale on the same controlled subset. To isolate these effects, we fix $m = 1$, $o_n = 1$, and $d = 1$ (i.e., a single non-nested operation) and vary entity size $n \in \{6, 8, 10, 12, 14\}$ and constraint count $k \in \{0, 1, 2, 3\}$. For each configuration, we generate 25 valid instances under the same 2000-second Cofola timeout, resulting in 2,000 problems.

As shown in Figure~\ref{fig:entity_constraint_analysis}, both factors consistently increase problem difficulty. 
Adding more constraints leads to a clear accuracy drop for most models, indicating that membership, 
positional, and relative-order constraints effectively disrupt simple counting patterns. Similarly, 
increasing $n$ enlarges the combinatorial search space and places greater pressure on numerical calculation and symbolic reasoning. 
These trends confirm that CombEval can control difficulty in a fine-grained and predictable way through structural generation parameters.

\begin{figure}[t]
    \centering
    \includegraphics[width=\linewidth]{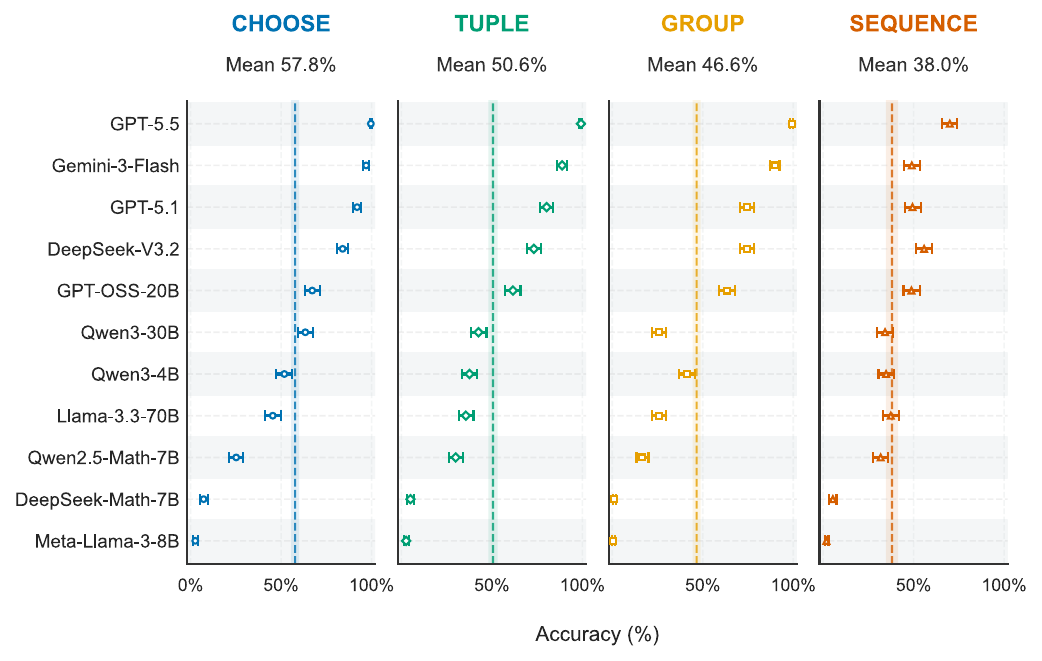}
    \caption{Accuracy across combinatorial operation types.}
    \label{fig:operation_type_results}
\end{figure}

Figure~\ref{fig:operation_type_results} further shows that operation type substantially affects model performance. 
Across models, \textsc{Sequence} is the most challenging category, with the lowest overall average accuracy, 
suggesting that ordered selections with relative-position constraints are particularly difficult for current LLMs, likely because they require more precise state tracking and constraint handling. 
\textsc{Choose} is comparatively easier, while \textsc{Group} and \textsc{Tuple} fall between these extremes. 
The same model hierarchy remains visible across operation types: stronger models such as \textbf{gpt\_5.5}, \textbf{gemini\_3\_flash\_preview\_thinking}, and \textbf{DeepSeek\_V3.2} consistently outperform smaller models, 
but their performance still varies substantially by operator. This confirms that CombEval not only separates model tiers overall, 
but also exposes fine-grained weaknesses tied to specific combinatorial operations.

\subsubsection{Impact of Reasoning Depth}

Beyond horizontal constraint accumulation, another key source of difficulty is multi-step operator nesting, 
which requires chain-structured reasoning. 
To study this factor in isolation, we use a minimally nested linear execution chain (e.g., $\textsc{Choose} \rightarrow \textsc{Choose} \rightarrow \textsc{Tuple}$). We eliminate confounding factors by fixing $n=10$, $m=1$, and $k=0$, while strictly enforcing that the nesting depth equals the number of operations ($d = o_n$). We systematically vary this depth across $\{1, 2, 3, 4\}$.
For each depth level, we generate 300 instances, resulting in 1,200 depth-controlled problems.

\begin{figure}[t]
    \centering
    \includegraphics[width=\linewidth]{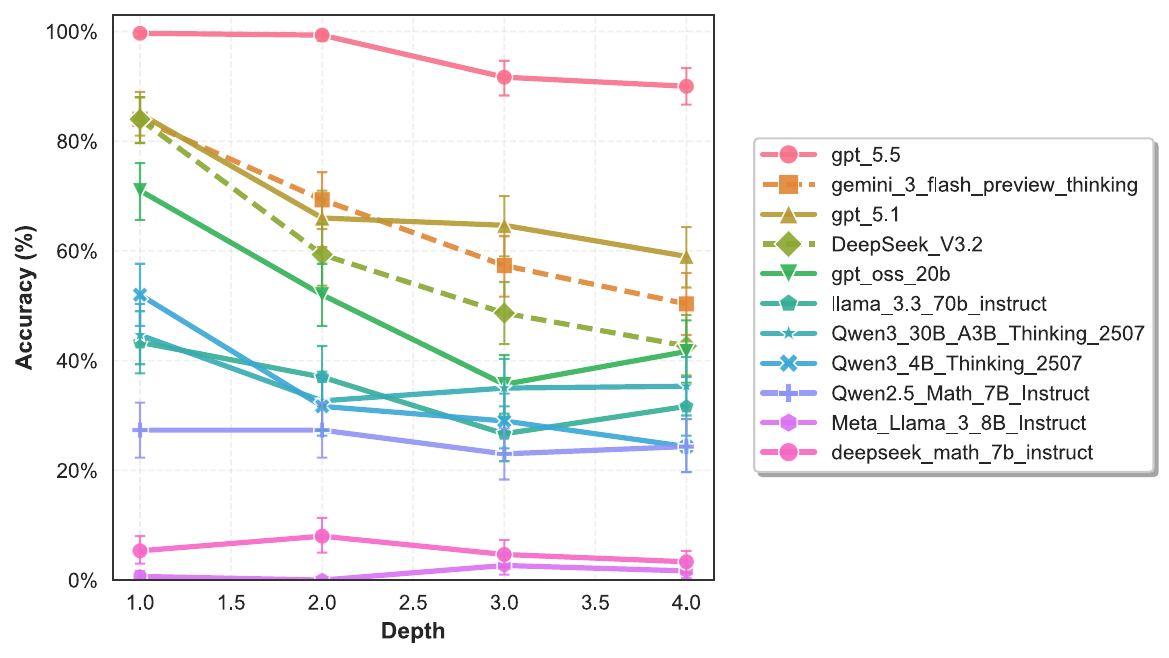}
    \caption{Impact of reasoning depth on model performance.}
    \label{fig:depth_analysis_exp_2}
\end{figure}

As illustrated in Figure~\ref{fig:depth_analysis_exp_2}, model accuracy degrades substantially as the nesting depth $d$ increases. Even in the absence of additional constraints and within a restricted entity space, each successive operation forces the model to track extended intermediate state chains, thereby exacerbating the risk of reasoning drift and error accumulation. This monotonic performance degradation quantitatively demonstrates CombEval's efficacy in modulating the difficulty of long-chain combinatorial reasoning via depth adjustments. Furthermore, while frontier models equipped with advanced reasoning paradigms (e.g., \textbf{DeepSeek-V3.2}, \textbf{gpt-5.5}, and \textbf{gemini-3-flash-preview-thinking}) exhibit greater resilience, they are not immune to this degradation.

\subsection{RQ3: Impact of Prompt Templates on LLM Reasoning}

A central question is whether LLMs fail on CombEval because they lack deep combinatorial reasoning ability, 
or because they are unfamiliar with our automatically generated custom templates. 
To isolate the influence of surface wording, we compare each original template problem with a solver-verified natural-language rewrite that preserves the same Cofola structure. 
The full rewrite-and-verification protocol is described in Appendix~\ref{app:RQ3PromptTemplateSetup}.
In total, we obtain 661 verified isomorphic pairs and randomly sample 300 pairs for this analysis. 
We report both single-prompt accuracy and paired consistency, where the latter requires the model to answer both logically equivalent formulations correctly.


\begin{table}[tbp]
\centering
\scriptsize
\setlength{\tabcolsep}{3pt}
\resizebox{\linewidth}{!}{%
\begin{tabular}{@{}llccc@{}}
\toprule
\textbf{Model} & \textbf{Setting}  & \textbf{Acc.} & $\Delta$ vs. Template & \textbf{95\% CI} \\
\midrule
\multirow{3}{*}{\textbf{DeepSeek-V3.2}} 
& Template & 80.42\% & -- & [77.41\%, 83.52\%] \\
& ReStory & 79.56\% & $-0.86$ & [76.41\%, 82.26\%] \\
& Both & 72.51\% & $-7.91$ & [68.89\%, 75.92\%] \\
\midrule
\multirow{3}{*}{\textbf{gpt-5.1}} 
& Template & 82.25\% & -- & [79.08\%, 85.59\%] \\
& ReStory & 80.25\% & $-2.00$ & [77.74\%, 83.26\%] \\
& Both & 74.23\% & $-8.02$ & [71.08\%, 77.85\%] \\
\midrule
\multirow{3}{*}{\textbf{\makecell{gemini-3-flash\\-preview-thinking}}} 
& Template & 87.38\% & -- & [84.74\%, 89.33\%] \\
& ReStory & 89.28\% & $+1.90$ & [87.67\%, 91.33\%] \\
& Both & 84.33\% & $-3.05$ & [81.48\%, 86.92\%] \\
\midrule
\multirow{3}{*}{\textbf{gpt-5.5}} 
& Template  & 92.59\% & -- & [89.89\%, 94.33\%] \\
& ReStory & 94.55\% & $+1.96$ & [92.74\%, 96.00\%] \\
& Both  & 91.83\% & $-0.76$ & [88.82\%, 93.67\%] \\
\bottomrule
\end{tabular}
}
\caption{Prompt-template robustness of \textbf{DeepSeek-V3.2}, \textbf{gemini-3-flash-preview-thinking}, \textbf{gpt-5.1}, and \textbf{gpt-5.5}. Template is the baseline; $\Delta$ reports percentage-point changes relative to Template. ReStory reports accuracy on rewritten prompts, and Both requires both formulations to be correct.}
\label{tab:consistency_analysis_gpt5}
\end{table}

As shown in Table~\ref{tab:consistency_analysis_gpt5}, all models perform similarly on original templates and rewritten prompts.
This suggests that CombEval's custom templates do not artificially depress model performance. 
However, paired consistency of relatively weaker models \textbf{DeepSeek-V3.2}, \textbf{gpt-5.1}, and \textbf{gemini-3-flash-preview-thinking} is lower than single-prompt accuracy (outside the confidence intervals of the original/rephrased template), indicating that some correct answers are not preserved under superficial rewrites.
This gap becomes negligible for \textbf{gpt-5.5}, dropping only $0.76\%$ compared to the original template.
Taken together, these results indicate that surface template variations can cause some performance fluctuations, especially for less capable (but still strong) models.
The latest stronger models, however, show greater robustness to these variations, suggesting that their reasoning capabilities are not tightly coupled to specific prompt formulations.

\subsection{Error Analysis and Qualitative Case Studies}

We manually analyze the $208$ wrong answers of \textbf{gpt-5.5}, of which $15$ are zero predictions and $193$ disagree on a non-zero value. Errors span all major operators including \textsc{Sequence}: $121$, \textsc{Group}: $62$, \textsc{Tuple}: $55$ and constraints \texttt{together}: $35$, \texttt{indexequalmember}: $34$, \texttt{lessthan}: $29$, \texttt{next\_to}: $27$ (note that one problem can have multiple operators and constraints, so these counts are not mutually exclusive).
They can be broadly categorized into: (i) collapsing a \texttt{together} block to one fixed internal order, missing its $k!$ permutations; (ii) reading ``\texttt{together(A) not in seq}'' as ``no two of $A$ adjacent'' instead of ``$A$ not contiguous''; (iii) treating multiple \texttt{choose\_tuple} on the same bag as one deal rather than independent draws; (iv) dropping the $(n{-}1)!$ quotient for multiset circles; (v) collapsing chains formed by \texttt{together} that share an entity; and (vi) collapsing internal ordering and direction of \texttt{together}$+$\texttt{next\_to} blocks in circles. 
Patterns (i)--(ii) reflect a semantic gap between natural-language and Cofola, while (iii)--(vi) are classical combinatorial slips. 
Representative examples are in \Cref{app:error_analysis_section}.

\section{Conclusion}

We present CombEval, an framework for dynamically generating diverse and difficulty-controllable combinatorial counting problems, used to evaluate the combinatorial counting capabilities of large language models.
Experimental results on CombEval reveal that current LLMs, including state-of-the-art closed-source models, still struggle with various combinatorial counting challenges, particularly those involving indistinguishable elements, complex sequencing constraints, and deep reasoning chains.
We anticipate that CombEval could serve as a valuable resource for advancing research in combinatorial counting and reasoning with LLMs, inspiring the development of more robust and capable models in this domain.
In future work, we plan to scale the LLM-facing suites to additional combinatorial object types, such as circular arrangements, partitions, graph objects, and numeric attributes, and to explore agentic techniques to enhance combinatorial reasoning performance.

\clearpage

\section*{Limitations}

While CombEval provides a structured benchmark for combinatorial reasoning, it has several limitations. 
The benchmark currently supports only English, and extending it to other languages would require additional template development and validation. 
In addition, verification with the Cofola solver is constrained by computational time and memory, which prevents the generation of more complex problem instances. 
Finally, human validation was performed only on a subset of samples, motivating future human-in-the-loop evaluation. 



\section*{Ethical considerations}

This work introduces CombEval, a benchmark for evaluating combinatorial counting capabilities of large language models. 
CombEval is intended for research purposes to advance understanding of LLM reasoning abilities. 
We do not foresee direct ethical concerns arising from the use of CombEval itself.

\section*{Acknowledgements}

Yuanhong Wang and Yuxu Zhou are supported by the National Natural Science Foundation of China (No.62506141). 
Ond\v{r}ej Ku\v{z}elka is supported by the Czech Science Foundation project ``The Automatic Combinatorialist'' (24-11820S). Additionally, this work is funded by the New Cornerstone Science Foundation via the XPLORER PRIZE.

\bibliography{custom}

@article{zhao2023survey,
	title        = {A survey of large language models},
	author       = {Zhao, Wayne Xin and Zhou, Kun and Li, Junyi and Tang, Tianyi and Wang, Xiaolei and Hou, Yupeng and Min, Yingqian and Zhang, Beichen and Zhang, Junjie and Dong, Zican and others},
	year         = 2023,
	journal      = {arXiv preprint arXiv:2303.18223},
	volume       = 1,
	number       = 2
}

@article{kalyan2024survey,
	title        = {A survey of GPT-3 family large language models including ChatGPT and GPT-4},
	author       = {Kalyan, Katikapalli Subramanyam},
	year         = 2024,
	journal      = {Natural Language Processing Journal},
	publisher    = {Elsevier},
	volume       = 6,
	pages        = 100048
}

@article{jiang2024survey,
	title        = {A survey on large language models for code generation},
	author       = {Jiang, Juyong and Wang, Fan and Shen, Jiasi and Kim, Sungju and Kim, Sunghun},
	year         = 2024,
	journal      = {arXiv preprint arXiv:2406.00515}
}

@article{xu2025towards,
	title        = {Towards large reasoning models: A survey of reinforced reasoning with large language models},
	author       = {Xu, Fengli and Hao, Qianyue and Zong, Zefang and Wang, Jingwei and Zhang, Yunke and Wang, Jingyi and Lan, Xiaochong and Gong, Jiahui and Ouyang, Tianjian and Meng, Fanjin and others},
	year         = 2025,
	journal      = {arXiv preprint arXiv:2501.09686}
}

@inproceedings{yang2025position,
	title        = {Position: Formal Mathematical Reasoning—A New Frontier in AI},
	author       = {Yang, Kaiyu and Poesia, Gabriel and He, Jingxuan and Li, Wenda and Lauter, Kristin E and Chaudhuri, Swarat and Song, Dawn},
	year         = 2025,
	booktitle    = {Forty-second International Conference on Machine Learning Position Paper Track}
}

@inproceedings{deng-etal-2024-investigating,
	title        = {Investigating Data Contamination in Modern Benchmarks for Large Language Models},
	author       = {Deng, Chunyuan  and Zhao, Yilun  and Tang, Xiangru  and Gerstein, Mark  and Cohan, Arman},
	year         = 2024,
	month        = jun,
	booktitle    = {Proceedings of the 2024 Conference of the North American Chapter of the Association for Computational Linguistics: Human Language Technologies (Volume 1: Long Papers)},
	publisher    = {Association for Computational Linguistics},
	address      = {Mexico City, Mexico},
	pages        = {8706--8719},
	doi          = {10.18653/v1/2024.naacl-long.482},
	url          = {https://aclanthology.org/2024.naacl-long.482/},
	editor       = {Duh, Kevin  and Gomez, Helena  and Bethard, Steven}
}

@article{kuang2025natural,
	title        = {Natural language understanding and inference with mllm in visual question answering: A survey},
	author       = {Kuang, Jiayi and Shen, Ying and Xie, Jingyou and Luo, Haohao and Xu, Zhe and Li, Ronghao and Li, Yinghui and Cheng, Xianfeng and Lin, Xika and Han, Yu},
	year         = 2025,
	journal      = {ACM Computing Surveys},
	publisher    = {ACM New York, NY},
	volume       = 57,
	number       = 8,
	pages        = {1--36}
}

@article{cobbe2021training,
	title        = {Training verifiers to solve math word problems},
	author       = {Cobbe, Karl and Kosaraju, Vineet and Bavarian, Mohammad and Chen, Mark and Jun, Heewoo and Kaiser, Lukasz and Plappert, Matthias and Tworek, Jerry and Hilton, Jacob and Nakano, Reiichiro and others},
	year         = 2021,
	journal      = {arXiv preprint arXiv:2110.14168}
}

@dataset{aime_1983_2024,
	title        = {AIME Problem Set 1983-2024},
	author       = {Hemish Veeraboina},
	year         = 2023,
	publisher    = {Kaggle},
	url          = {https://www.kaggle.com/datasets/hemishveeraboina/aime-problem-set-1983-2024}
}

@article{zheng2021minif2f,
	title        = {Minif2f: a cross-system benchmark for formal olympiad-level mathematics},
	author       = {Zheng, Kunhao and Han, Jesse Michael and Polu, Stanislas},
	year         = 2021,
	journal      = {arXiv preprint arXiv:2109.00110}
}

@article{xuejun2025mathesis,
	title        = {Mathesis: Towards Formal Theorem Proving from Natural Languages},
	author       = {Xuejun, Yu and Zhong, Jianyuan and Feng, Zijin and Zhai, Pengyi and Yousefzadeh, Roozbeh and Ng, Wei Chong and Liu, Haoxiong and Shou, Ziyi and Xiong, Jing and Zhou, Yudong and others},
	year         = 2025,
	journal      = {arXiv preprint arXiv:2506.07047}
}

@article{liu2025combibench,
	title        = {CombiBench: Benchmarking LLM capability for combinatorial mathematics},
	author       = {Liu, Junqi and Lin, Xiaohan and Bayer, Jonas and Dillies, Yael and Jiang, Weijie and Liang, Xiaodan and Soletskyi, Roman and Wang, Haiming and Xie, Yunzhou and Xiong, Beibei and others},
	year         = 2025,
	journal      = {arXiv preprint arXiv:2505.03171}
}

@article{valiant_complexity_1979,
	title        = {The {{Complexity}} of {{Enumeration}} and {{Reliability Problems}}},
	author       = {Valiant, Leslie G.},
	year         = 1979,
	journal      = {SIAM Journal on Computing},
	volume       = 8,
	number       = 3,
	pages        = {410--421},
	issn         = {0097-5397, 1095-7111}
}

@inproceedings{ariyani2025there,
	title        = {There’s no such thing as simple reasoning for LLMs},
	author       = {Ariyani, Nurul Fajrin and Bouraoui, Zied and Booth, Richard and Schockaert, Steven},
	year         = 2025,
	booktitle    = {Findings of the Association for Computational Linguistics: ACL 2025},
	pages        = {4503--4514}
}

@article{saparov2023testing,
	title        = {Testing the general deductive reasoning capacity of large language models using ood examples},
	author       = {Saparov, Abulhair and Pang, Richard Yuanzhe and Padmakumar, Vishakh and Joshi, Nitish and Kazemi, Mehran and Kim, Najoung and He, He},
	year         = 2023,
	journal      = {Advances in Neural Information Processing Systems},
	volume       = 36,
	pages        = {3083--3105}
}

@inproceedings{wan2024logicasker,
	title        = {LogicAsker: Evaluating and improving the logical reasoning ability of large language models},
	author       = {Wan, Yuxuan and Wang, Wenxuan and Yang, Yiliu and Yuan, Youliang and Huang, Jen-tse and He, Pinjia and Jiao, Wenxiang and Lyu, Michael},
	year         = 2024,
	booktitle    = {Proceedings of the 2024 Conference on Empirical Methods in Natural Language Processing},
	pages        = {2124--2155}
}

@article{totis_lifted_2023,
	title        = {Lifted {{Reasoning}} for {{Combinatorial Counting}}},
	author       = {Totis, Pietro and Davis, Jesse and De Raedt, Luc and Kimmig, Angelika},
	year         = 2023,
	month        = jan,
	journal      = {Journal of Artificial Intelligence Research},
	volume       = 76,
	pages        = {1--58},
	doi          = {10.1613/jair.1.14062},
	issn         = {1076-9757}
}

@inproceedings{pak_complexity_2019,
	title        = {{COMPLEXITY PROBLEMS IN ENUMERATIVE COMBINATORICS}},
	author       = {Pak, {\relax Igor}},
	year         = 2019,
	month        = may,
	booktitle    = {Proceedings of the {{International Congress}} of {{Mathematicians}} ({{ICM}} 2018)},
	publisher    = {WORLD SCIENTIFIC},
	address      = {Rio de Janeiro, Brazil},
	doi          = {10.1142/9789813272880_0176},
	eprint       = {1803.06636},
	primaryclass = {cs, math}
}

@inproceedings{shi2022stepgame,
	title        = {Stepgame: A new benchmark for robust multi-hop spatial reasoning in texts},
	author       = {Shi, Zhengxiang and Zhang, Qiang and Lipani, Aldo},
	year         = 2022,
	booktitle    = {Proceedings of the AAAI conference on artificial intelligence},
	volume       = 36,
	pages        = {11321--11329}
}

@inproceedings{wang2025generating,
	title        = {Generating pedagogically meaningful visuals for math word problems: A new benchmark and analysis of text-to-image models},
	author       = {Wang, Junling and Rutkiewicz, Anna and Wang, April and Sachan, Mrinmaya},
	year         = 2025,
	booktitle    = {Findings of the Association for Computational Linguistics: ACL 2025},
	pages        = {11229--11257}
}

@article{stanley2011enumerative,
	title        = {Enumerative combinatorics volume 1 second edition},
	author       = {Stanley, Richard P},
	year         = 2011,
	journal      = {Cambridge studies in advanced mathematics}
}

@article{ferraris2015counting,
	title        = {Counting sub-multisets of fixed cardinality},
	author       = {Ferraris, Sebastiano and Mendelson, Alex and Ballesio, Gerardo and Vercauteren, Tom},
	year         = 2015,
	journal      = {arXiv preprint arXiv:1511.06142}
}

@inproceedings{hendrycks2021measuring,
	title        = {Measuring Mathematical Problem Solving With the MATH Dataset},
	author       = {Hendrycks, Dan and Burns, Collin and Kadavath, Saurav and Arora, Akul and Basart, Steven and Tang, Eric and Song, Dawn and Steinhardt, Jacob},
	year         = 2021,
	booktitle    = {Proceedings of NeurIPS}
}

@article{gao2023retrieval,
	title        = {Retrieval-augmented generation for large language models: A survey},
	author       = {Gao, Yunfan and Xiong, Yun and Gao, Xinyu and Jia, Kangxiang and Pan, Jinliu and Bi, Yuxi and Dai, Yixin and Sun, Jiawei and Wang, Haofen and Wang, Haofen},
	year         = 2023,
	journal      = {arXiv preprint arXiv:2312.10997},
	volume       = 2,
	number       = 1
}

@article{balloccu2024leak,
	title        = {Leak, cheat, repeat: Data contamination and evaluation malpractices in closed-source LLMs},
	author       = {Balloccu, Simone and Schmidtov{\'a}, Patr{\'\i}cia and Lango, Mateusz and Du{\v{s}}ek, Ond{\v{r}}ej},
	year         = 2024,
	journal      = {arXiv preprint arXiv:2402.03927}
}

@article{golchin2023time,
	title        = {Time travel in llms: Tracing data contamination in large language models},
	author       = {Golchin, Shahriar and Surdeanu, Mihai},
	year         = 2023,
	journal      = {arXiv preprint arXiv:2308.08493}
}

@article{opedal2024mathgap,
	title        = {Mathgap: Out-of-distribution evaluation on problems with arbitrarily complex proofs},
	author       = {Opedal, Andreas and Shirakami, Haruki and Sch{\"o}lkopf, Bernhard and Saparov, Abulhair and Sachan, Mrinmaya},
	year         = 2024,
	journal      = {arXiv preprint arXiv:2410.13502}
}

@article{zhou2025lessleak,
	title        = {Lessleak-bench: A first investigation of data leakage in llms across 83 software engineering benchmarks},
	author       = {Zhou, Xin and Weyssow, Martin and Widyasari, Ratnadira and Zhang, Ting and He, Junda and Lyu, Yunbo and Chang, Jianming and Zhang, Beiqi and Huang, Dan and Lo, David},
	year         = 2025,
	journal      = {arXiv preprint arXiv:2502.06215}
}

@article{balloccu2024benchmark,
  title={Benchmark data contamination of large language models: A survey, 2024},
  author={Xu, Cheng and Guan, Shuhao and Greene, Derek and Kechadi, M-Tahar},
  journal={URL https://arxiv. org/abs/2406.04244}
}

@inproceedings{gao2023pal,
	title        = {Pal: Program-aided language models},
	author       = {Gao, Luyu and Madaan, Aman and Zhou, Shuyan and Alon, Uri and Liu, Pengfei and Yang, Yiming and Callan, Jamie and Neubig, Graham},
	year         = 2023,
	booktitle    = {International Conference on Machine Learning},
	pages        = {10764--10799},
	organization = {PMLR}
}

@article{mirzadeh2024gsm,
	title        = {Gsm-symbolic: Understanding the limitations of mathematical reasoning in large language models},
	author       = {Mirzadeh, Iman and Alizadeh, Keivan and Shahrokhi, Hooman and Tuzel, Oncel and Bengio, Samy and Farajtabar, Mehrdad},
	year         = 2024,
	journal      = {arXiv preprint arXiv:2410.05229}
}

@article{lai2025making,
	title        = {Making Mathematical Reasoning Adaptive},
	author       = {Lai, Zhejian and Geng, Xiang and Wang, Zhijun and Bai, Yang and Li, Jiahuan and Weng, Rongxiang and Wang, Jingang and Cao, Xuezhi and Cai, Xunliang and Huang, Shujian},
	year         = 2025,
	journal      = {arXiv preprint arXiv:2510.04617}
}

@article{boye2025large,
	title        = {Large language models and mathematical reasoning failures},
	author       = {Boye, Johan and Moell, Birger},
	year         = 2025,
	journal      = {arXiv preprint arXiv:2502.11574}
}

@article{shrestha2025mathematical,
	title        = {Mathematical reasoning in large language models: Assessing logical and arithmetic errors across wide numerical ranges},
	author       = {Shrestha, Safal and Kim, Minwu and Ross, Keith},
	year         = 2025,
	journal      = {arXiv preprint arXiv:2502.08680}
}

@misc{DeepSeek-Math-7B-Instruction,
	title        = {DeepSeekMath: Pushing the Limits of Mathematical Reasoning in Open Language Models},
	author       = {Zhihong Shao and Peiyi Wang and Qihao Zhu and Runxin Xu and Junxiao Song and Xiao Bi and Haowei Zhang and Mingchuan Zhang and Y. K. Li and Y. Wu and Daya Guo},
	year         = 2024,
	url          = {https://arxiv.org/abs/2402.03300},
	eprint       = {2402.03300},
	archiveprefix = {arXiv},
	primaryclass = {cs.CL}
}

@article{Qwen2.5-Math-7B-Instruct,
	title        = {Qwen2.5-Math Technical Report: Toward Mathematical Expert Model via Self-Improvement},
	author       = {An Yang and Beichen Zhang and Binyuan Hui and Bofei Gao and Bowen Yu and Chengpeng Li and Dayiheng Liu and Jianhong Tu and Jingren Zhou and Junyang Lin and Keming Lu and Mingfeng Xue and Runji Lin and Tianyu Liu and Xingzhang Ren and Zhenru Zhang},
	year         = 2024,
	journal      = {arXiv preprint arXiv:2409.12122}
}

@inproceedings{van_den_broeck_lifted_2011,
	title        = {Lifted {{Probabilistic Inference}} by {{First-Order Knowledge Compilation}}},
	author       = {Van Den Broeck, Guy and Taghipour, Nima and Meert, Wannes and Davis, Jesse and De Raedt, Luc},
	year         = 2011,
	booktitle    = {{{IJCAI}} 2011, {{Proceedings}} of the 22nd {{International Joint Conference}} on {{Artificial Intelligence}}, {{Barcelona}}, {{Catalonia}}, {{Spain}}, {{July}} 16-22, 2011},
	series       = {{{IJCAI}} 2011},
	volume       = 29,
	pages        = {2178--2185},
	doi          = {10.5591/978-1-57735-516-8/IJCAI11-363},
	isbn         = {978-1-57735-512-0},
	eprint       = {1412.0315}
}

@inproceedings{thurley2006sharpsat,
	title        = {sharpSAT--counting models with advanced component caching and implicit BCP},
	author       = {Thurley, Marc},
	year         = 2006,
	booktitle    = {International Conference on Theory and Applications of Satisfiability Testing},
	pages        = {424--429},
	organization = {Springer}
}

@book{gebser2022answer,
	title        = {Answer set solving in practice},
	author       = {Gebser, Martin and Kaminski, Roland and Kaufmann, Benjamin and Schaub, Torsten},
	year         = 2022,
	publisher    = {Springer Nature}
}

@misc{Qwen3,
	title        = {Qwen3 Technical Report},
	author       = {{Qwen Team}},
	year         = 2025,
	url          = {https://arxiv.org/abs/2505.09388},
	eprint       = {2505.09388},
	archiveprefix = {arXiv},
	primaryclass = {cs.CL}
}

@misc{openai2025gptoss120bgptoss20bmodel,
	title        = {gpt-oss-120b \& gpt-oss-20b Model Card},
	author       = {OpenAI},
	year         = 2025,
	url          = {https://arxiv.org/abs/2508.10925},
	eprint       = {2508.10925},
	archiveprefix = {arXiv},
	primaryclass = {cs.CL}
}

@article{frisch2008essence,
	title        = {Essence: A constraint language for specifying combinatorial problems},
	author       = {Frisch, Alan M and Harvey, Warwick and Jefferson, Chris and Mart{\'\i}nez-Hern{\'a}ndez, Bernadette and Miguel, Ian},
	year         = 2008,
	journal      = {Constraints},
	publisher    = {Springer},
	volume       = 13,
	number       = 3,
	pages        = {268--306}
}

@article{akgun2022conjure,
	title        = {Conjure: Automatic generation of constraint models from problem specifications},
	author       = {Akg{\"u}n, {\"O}zg{\"u}r and Frisch, Alan M and Gent, Ian P and Jefferson, Christopher and Miguel, Ian and Nightingale, Peter},
	year         = 2022,
	journal      = {Artificial Intelligence},
	publisher    = {Elsevier},
	volume       = 310,
	pages        = 103751
}

@inproceedings{gent2006minion,
	title        = {MINION: A Fast, Scalable, Constraint Solver$^1$},
	author       = {Gent, Ian P and Jefferson, Chris},
	year         = 2006,
	booktitle    = {ECAI 2006: 17th European Conference on Artificial Intelligence, August 29-September 1, 2006, Riva del Garda, Italy: including Prestigious Applications of Intelligent Systems (PAIS 2006): proceedings},
	volume       = 141,
	pages        = 98,
	organization = {Ios Press}
}

@misc{deepseekai2025deepseekv32,
	title        = {DeepSeek-V3.2: Pushing the Frontier of Open Large Language Models},
	author       = {DeepSeek-AI},
	year         = 2025
}

@inproceedings{taghipour2012lifted,
	title        = {Lifted variable elimination with arbitrary constraints},
	author       = {Taghipour, Nima and Fierens, Daan and Davis, Jesse and Blockeel, Hendrik},
	year         = 2012,
	booktitle    = {Artificial Intelligence and Statistics},
	pages        = {1194--1202},
	organization = {PMLR}
}

@article{dubey2024llama,
	title        = {The llama 3 herd of models},
	author       = {Dubey, Abhimanyu and Jauhri, Abhinav and Pandey, Abhinav and Kadian, Abhishek and Al-Dahle, Ahmad and Letman, Aiesha and Mathur, Akhil and Schelten, Alan and Yang, Amy and Fan, Angela and others},
	year         = 2024,
	journal      = {arXiv preprint arXiv:2407.21783}
}

@inproceedings{DBLP:conf/uai/BremenK21,
	title        = {Faster lifting for two-variable logic using cell graphs},
	author       = {Timothy van Bremen and Ondrej Kuzelka},
	year         = 2021,
	booktitle    = {Proceedings of the Thirty-Seventh Conference on Uncertainty in Artificial Intelligence, {UAI} 2021, Virtual Event, 27-30 July 2021},
	publisher    = {{AUAI} Press},
	series       = {Proceedings of Machine Learning Research},
	volume       = 161,
	pages        = {1393--1402},
	url          = {https://proceedings.mlr.press/v161/bremen21a.html},
	editor       = {Cassio P. de Campos and Marloes H. Maathuis and Erik Quaeghebeur},
	bibsource    = {dblp computer science bibliography, https://dblp.org}
}

@article{kuzelka_weighted_2021,
	title        = {Weighted First-Order Model Counting in the Two-Variable Fragment with Counting Quantifiers},
	author       = {Kuzelka, Ondrej},
	year         = 2021,
	month        = mar,
	journal      = {Journal of Artificial Intelligence Research},
	volume       = 70,
	pages        = {1281--1307},
	doi          = {10.1613/jair.1.12320},
	issn         = {1076-9757},
	eprint       = {2007.05619}
}

@article{toth_lifted_2022-1,
	title        = {Lifted Inference with Linear Order Axiom},
	author       = {T{\'o}th, Jan and Ku{\v z}elka, Ond{\v r}ej},
	year         = 2022,
	journal      = {Proceedings of the AAAI Conference on Artificial Intelligence},
	volume       = 37,
	number       = 10,
	pages        = {12295--12304},
	doi          = {10.1609/aaai.v37i10.26449}
}

@article{DBLP:journals/corr/abs-2507-19182,
	title        = {Faster Lifting for Ordered Domains with Predecessor Relations},
	author       = {Kuncheng Zou and Jiahao Mai and Yonggang Zhang and Yuyi Wang and Ondrej Kuzelka and Yuanhong Wang and Yi Chang},
	year         = 2025,
	journal      = {CoRR},
	volume       = {abs/2507.19182},
	doi          = {10.48550/ARXIV.2507.19182},
	url          = {https://doi.org/10.48550/arXiv.2507.19182},
	eprinttype   = {arXiv},
	eprint       = {2507.19182},
	bibsource    = {dblp computer science bibliography, https://dblp.org}
}

@misc{wang2026solvingcombinatorialcountingproblems,
      title={Solving Combinatorial Counting Problems with Weighted First-Order Model Counting}, 
      author={Yuanhong Wang and Juhua Pu and Yuxu Zhou and Yuyi Wang and Ondřej Kuželka},
      year={2026},
      eprint={2605.24845},
      archivePrefix={arXiv},
      primaryClass={cs.AI},
      url={https://arxiv.org/abs/2605.24845}, 
}

\clearpage

\appendix

\label{sec:appendix}

\section{Problem Examples}

\paragraph{\textsc{Choose}} Set $set_0$ contains: $e_4$, $e_1$, $e_5$, $e_3$, $e_2$, $e_6$, $e_7$, $e_8$. From $set_0$, we repeatedly select 16 elements (with replacement) to form $choose_0$. $choose_0$ contains $e_4$. The element $e_5$ must appear at most 11 times in $choose_0$. $choose_0$ is disjoint from $\{e_1, e_2, e_6\}$. How many different ways can we obtain $choose_0$? Count each distinct combination of $choose_0$ as one way. (Answer: 3841)

\paragraph{\textsc{Group}} Set $set_0$ contains: $e_6$, $e_2$, $e_9$, $e_{10}$, $e_3$, $e_7$, $e_4$, $e_8$, $e_5$, $e_1$. We divide $set_0$ into 4 groups (allowing empty groups and groups are labeled), storing the result in $group_0$. The 4th division group of $group_0$ is a subset of $\{e_1, e_{10}, e_2, e_3, e_5, e_6, e_8\}$. The 3rd division group of $group_0$ is disjoint from $\{e_8\}$. How many different ways can we obtain $group_0$? Count each distinct combination of $group_0$ as one way. (Answer: 331776)

\paragraph{\textsc{Sequence}} Set $set_0$ contains: $e_5$, $e_1$, $e_{11}$, $e_3$, $e_2$, $e_8$, $e_{10}$, $e_4$, $e_{12}$, $e_6$, $e_9$, $e_7$, $e_{13}$, $e_{14}$. We draw 4 elements from $set\_0$ and arrange them in sequence $choose\_sequence_0$. Sequence $choose\_sequence_0$ must contain $e_8$ and $e_{10}$ such that $e_8$ is immediately before $e_{10}$. How many different ways can we obtain $choose\_sequence_0$? Count each distinct combination of $choose\_sequence_0$ as one way. (Answer: 396)

\paragraph{\textsc{Tuple}} Set $set_0$ contains: $e_3$, $e_2$, $e_4$, $e_6$, $e_1$, $e_8$, $e_5$, $e_7$. We draw 4 elements from $set_0$ (with replacement) and arrange them in $choose\_tuple_0$. The element at index 4 of $choose\_tuple_0$ must not be $e_4$. The element $e_5$ must appear at most 3 times in $choose\_tuple_0$. The element at index 2 of $choose\_tuple_0$ must be $e_6$. How many different ways can we obtain $choose\_tuple_0$? Count each distinct combination of $choose\_tuple_0$ as one way. (Answer: 448)

\section{Solvers for Combinatorial Counting}
\label{app:Existing_Solvers_for_Combinatorial_Counting}

We briefly review the existing solvers for combinatorial counting (CO) problems. 
These algorithmic approaches can be broadly categorized into Constraint Satisfaction Problems (CSPs), propositional logic frameworks (including Answer Set Programming and \#SAT), and lifted inference methods.

\paragraph{Constraint Satisfaction Problems (CSPs)}

Constraint-based frameworks typically employ a two-step approach: modeling and solving. 
In modeling, problems are defined using high-level constraint languages, such as ESSENCE \cite{frisch2008essence,akgun2022conjure}, which allow users to specify combinatorial objects using primitives like sets, multisets, and partitions, along with constraints on valid configurations. 
Then, these high-level models are refined into lower-level representations (e.g., ESSENCE$'$) and compiled into a counting constraint satisfaction problem (\#CSP). 
Finally, specialized solvers (e.g., Minion \cite{gent2006minion}) are used to count the number of solutions satisfying all constraints, answering the original combinatorial counting question.

\paragraph{Propositional Logic and Answer Set Programming (ASP)}

Logic-based solvers approach counting by finding the number of "models" (satisfying truth assignments) for a given logical theory, i.e., via $\#\SAT$ \cite{thurley2006sharpsat} or $\#count\ \{l_1, l_2, \dots, l_n\}$ queries (which count the number of true literals among $l_1, l_2, \dots, l_n$) \cite{gebser2022answer}.
The main limitation of these methods is that they do not natively support high-level combinatorial constructs, e.g., sets or multisets, requiring explicit encoding of such structures into propositional variables and clauses.

\paragraph{Lifted Inference Methods}

Lifted inference frameworks, such as Forclift \cite{van_den_broeck_lifted_2011}, GC-FOVE \cite{taghipour2012lifted}, WFOMC \cite{kuzelka_weighted_2021,DBLP:conf/uai/BremenK21,toth_lifted_2022-1}, aim to count solutions without exhaustively enumerating every individual state. 
These methods rely on the principle of exchangeability (symmetries of the domain constants), reasoning about groups of individuals together rather than unique constants. 
This is very similar to how combinatorial counting formulas operate on groups of indistinguishable objects, making lifted inference a natural fit for CO problems. 
Recently, \citet{totis_lifted_2023} attempted to bridge lifted inference and \#CSP, proposing a lifted reasoning framework specifically designed for combinatorial counting tasks. 
This framework, containing a Domain-Specific Language (DSL) called \emph{CoLa} and the solver \emph{CoSo}, showed promising results on modeling and solving various CO problems \cite{totis_lifted_2023}.
Cofola~\citep{wang2026solvingcombinatorialcountingproblems} used in this work extends this direction with a typed object language and a WFOMC compilation pipeline, supporting multiple dependent objects, sets and bags, tuples, sequences, circles, partitions, compositions, and relative or group-level constraints.



\begin{figure}[tbp]
    \centering
    \begin{tcolorbox}[
        colback=gray!5,
        colframe=black!70,
        boxrule=0.8pt,
        arc=2mm,
        title=\textbf{Evaluation Prompt for Zero-Shot Reasoning},
        fonttitle=\bfseries\sffamily,
        left=6pt, right=6pt, top=6pt, bottom=6pt
    ]
    \ttfamily\small
    You are an expert in combinatorics. Your task is to solve combinatorics problems by providing a step-by-step reasoning process followed by a single, exact numerical answer.
    
    \vspace{0.5em}
    \noindent Instructions:
    \begin{enumerate}[leftmargin=*, nosep, label=\arabic*.]
        \item First, reason step-by-step to solve the problem. Show all intermediate steps, including the use of standard combinatorial notation and formulas.
        \item Compute the exact numerical value; do not leave expressions unevaluated.
        \item After completing the reasoning, report the final answer as a single number enclosed in \textbf{\textbackslash boxed\{\}}.
        \item Do not include any text after the boxed answer.
    \end{enumerate}
    
    \vspace{0.5em}
    \noindent Question:\\
    \{question\}
    \end{tcolorbox}
    \caption{The prompt template used during evaluation. The placeholder \texttt{\{question\}} is instantiated with a specific problem instance at inference time.}
    \label{fig:evaluation_prompt}

\end{figure}
\begin{figure}[tbp]
    \centering
    \begin{tcolorbox}[
        colback=gray!5,
        colframe=black!70,
        boxrule=0.8pt,
        arc=2mm,
        title=\textbf{Evaluation Prompt for Code Augmented Reasoning},
        fonttitle=\bfseries\sffamily,
        left=6pt, right=6pt, top=6pt, bottom=6pt
    ]
    \ttfamily\small
    You are an expert in combinatorics and Python programming. Your task is to solve combinatorics problems by writing Python code.
    
    \vspace{0.5em}
    \noindent Instructions:
    \begin{enumerate}[leftmargin=*, nosep, label=\arabic*.]
        \item You must ONLY output executable Python code. Do not include any natural language explanations, introduction, text, or mathematical derivation.
        \item The code must be self-contained and calculate the exact numerical answer to the problem.
        \item Print or return the final result within the code. Do not add any text or formatting (like \textbackslash boxed\{\}) outside the code block.
    \end{enumerate}
    
    \vspace{0.5em}
    \noindent Question:\\
    \{question\}
    \end{tcolorbox}
    \caption{The prompt template used during code-augmented evaluation (Python setting). The placeholder \texttt{\{question\}} is instantiated with a specific problem instance at inference time.}
    \label{fig:code_evaluation_prompt}
\end{figure}

\section{Experimental Details}
\label{app:experimental_details}

\subsection{Capabilities of LLMs in Combinatorial Reasoning dataset distribution}
\label{app:datasetdistribution}

For RQ1 (Capabilities of LLMs in Combinatorial Reasoning), we construct an
evaluation dataset across the four operation groups sampled in the main
CombEval suite: \textsc{Choose}, \textsc{Group}, \textsc{Sequence}, and
\textsc{Tuple}.
The final dataset comprises 1,500 problem instances in total.
By operation type, instances are distributed as follows: \textsc{Tuple}
(589, 39.3\%), \textsc{Group} (337, 22.5\%), \textsc{Sequence} (297,
19.8\%), and \textsc{Choose} (277, 18.5\%).

Instance difficulty varies across multiple axes.
By reasoning depth, most instances involve a single operator application
(depth-1: 1,272, 84.8\%), with decreasing numbers at deeper nesting levels
(depth-2: 193, 12.9\%; depth-3: 32, 2.1\%; depth-4: 3, 0.2\%).
By constraint count, 574 instances (38.3\%) contain no constraints, 413
(27.5\%) contain one constraint, 303 (20.2\%) contain two constraints, and
210 (14.0\%) contain three constraints.
By operator count, the majority of problems use a single combinatorial
operator (858, 57.2\%), with 441 (29.4\%) using two operators, 146 (9.7\%)
using three operators, and 55 (3.7\%) using four operators.
All instances are verified against the Cofola solver to ensure ground-truth
answers exist.

\begin{figure*}[t]
    \centering
    \includegraphics[width=\linewidth]{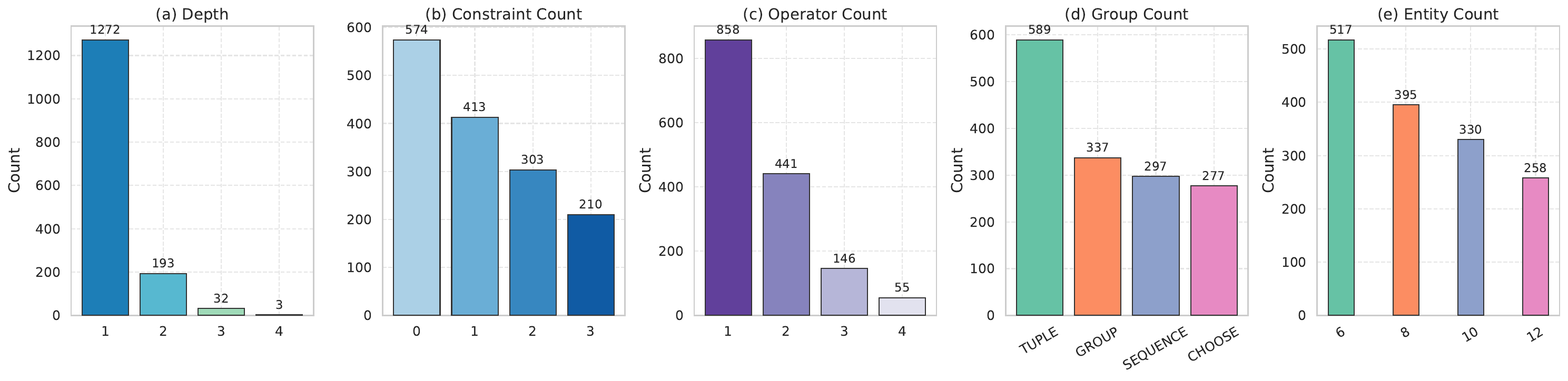}
    \caption{Distribution of problem instances across operation types and
    difficulty parameters for the RQ1 dataset.}
    \label{fig:distributionforRQ1Dataset}
\end{figure*}

\subsection{Evaluation Prompts}
\label{app:Evaluation_Prompts}
All models are evaluated using a standardized prompt template designed to elicit step-by-step chain-of-thought reasoning followed by a clearly marked final answer. 
This prompt engineering approach serves two purposes: 
(1) it encourages models to show their reasoning process, enabling error analysis and identification of failure modes, and 
(2) it enforces a consistent output format that facilitates automatic answer extraction and evaluation.

Figure~\ref{fig:evaluation_prompt} presents the complete prompt template used across all experiments. 
The prompt explicitly instructs models to act as combinatorics experts, 
decompose problems into intermediate steps using standard notation (e.g., $\binom{n}{k}$, permutations, factorial expressions), 
and compute exact numerical values rather than leaving symbolic expressions unevaluated. 
Critically, the final answer must be enclosed in \texttt{\textbackslash boxed\{\}} delimiters with no additional text afterward, allowing reliable extraction via regex pattern matching.

During evaluation, the \texttt{\{question\}} placeholder is replaced with individual problem instances from the dataset. 
We apply this template uniformly to all problem types (\textsc{Choose}, \textsc{Group}, \textsc{Sequence}, \textsc{Tuple}) without type-specific modifications, ensuring fair comparison across tasks. 
Models receive no few-shot examples or additional context beyond the problem statement itself, constituting a pure zero-shot evaluation setting.

\subsection{Runtime as Difficulty Indicator}
\label{app:Runtime_as_Difficulty_Indicator}
\begin{figure*}[htbp]
  \centering
  \includegraphics[width=\linewidth]{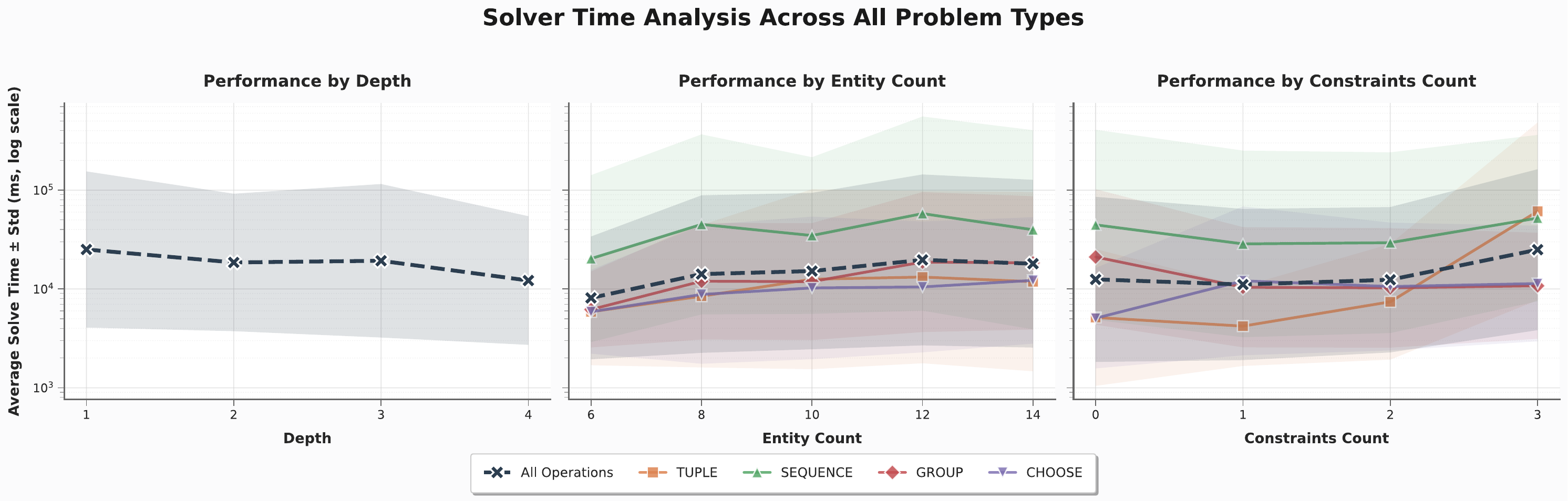}
  \caption{
    Performance analysis of solver runtime across different problem parameters. 
  }
  \label{fig:solver_time_analysis}
\end{figure*}

To validate that generated problems exhibit genuine computational complexity, we analyze the relationship between problem parameters and solver runtime, which serves as an objective measure of intrinsic difficulty independent of model-specific biases.The scaling of solver performance with depth, entity size, and constraint count on a logarithmic scale is shown in Figure~\ref{fig:solver_time_analysis}. 

The results reveal several important patterns that support our difficulty parameterization. Notably, solver runtime exhibits a marginal decrease as the reasoning depth scales from 1 to 4. This counterintuitive behavior stems from our isolated experimental design for depth evaluation, where the constraint count is strictly fixed to $k=0$ to eliminate confounding factors. Without the overhead of processing complex constraint interactions, the solver can efficiently traverse the nested chain-structured operators (e.g., $\textsc{Choose} \rightarrow \textsc{Sequence} \rightarrow \dots$), bypassing intensive filtering or backtracking. Consequently, deeper nesting under zero-constraint settings effectively simplifies the solver's localized search paths, resulting in a slight reduction in execution time. In contrast, entity size and constraint count lead to substantial, exponential increases in solver runtime, reflecting their dominant impact on expanding the combinatorial search space and complicating constraint satisfaction.

\section{Experimental Details for RQ3}
\label{app:RQ3PromptTemplateSetup}
\subsection{Dataset}
We construct the evaluation dataset for RQ3 directly from the problem instances used in our main experiments (RQ1 and RQ2).
Each problem is originally synthesized by the CombEval framework in the formal Cofola representation, then automatically translated into natural language using a structured template, producing a baseline natural language formulation.
To obtain a linguistically diversified variant while strictly preserving semantic equivalence, we apply a closed-loop verification pipeline to rewrite each problem into a MATH-style natural language formulation, as detailed in the next subsection.
As shown in Figure~\ref{fig:problem-examples}, the same underlying combinatorial structure can be expressed in both the template-based and rewritten natural-language forms while maintaining semantic equivalence.

\subsection{Closed-Loop Verification for Semantic Preservation}
A core challenge in evaluating prompt sensitivity is ensuring that rewritten problems are mathematically isomorphic to the originals---i.e., any observed performance variation must be attributable solely to surface-level linguistic changes, not to unintended semantic drift. 
To this end, we design a \textbf{closed-loop verification} pipeline comprising three stages: Style Transfer, Formal Back-translation, and Solver Verification.

\paragraph{Step 1: Style Transfer.}
We leverage \textbf{gemini-3-flash-preview-thinking} to rewrite each template-generated problem into a MATH-style natural language formulation.
The rewriting is conducted in a retrieval-augmented generation (RAG) setting~\cite{gao2023retrieval}: reference examples are retrieved from the ``Counting and Probability'' category of the MATH dataset~\cite{hendrycks2021measuring} (Figure~\ref{fig:math_examples}) and provided as few-shot stylistic guidance.
The rewriting prompt (Figure~\ref{fig:rewrite_prompt}) explicitly conditions generation on the underlying Cofola representation,separating mathematical logic from surface-level language to encourage diverse narrative reformulations without altering the solution space.
\paragraph{Step 2: Formal Back-translation.}To guard against condition omission or semantic shift introduced during rewriting, we employ the same LLM to translate the rewritten problem \textit{back} into Cofola code using a dedicated NL-to-Cofola translation prompt (Figure~\ref{fig:nl_to_cofola_prompt}).
This prompt provides exhaustive one-to-one mappings between natural language patterns and Cofola constructs, enforcing a deterministic, line-by-line translation that recovers the formal structure from the paraphrased text.
\paragraph{Step 3: Solver Verification.}The back-translated Cofola expression is fed into the Cofola constraint solver to compute its exact solution.If and only if the solver's output matches the ground-truth answer of the original problem exactly---across all valid combinatorial configurations---is the rewritten instance accepted as a semantically lossless paraphrase.Instances that fail this verification are discarded and re-generated (with different random seeds for the style transfer stage) until convergence or a maximum retry budget is exhausted.
This closed-loop protocol guarantees that every accepted Rewritten variant encodes precisely the same combinatorial enumeration problem as its Math and Cards counterparts,enabling a rigorous, confounding-free evaluation of model robustness to paraphrasing in RQ3.

\subsection{Prompt for Problem Rewriting}
\begin{figure*}[ht]
\centering
\begin{tcolorbox}[
colback=gray!5,
colframe=black!70,
boxrule=0.8pt,
arc=2mm,
title=\textbf{Prompt Template for Problem Rewriting},
fonttitle=\bfseries\sffamily,
fontupper=\ttfamily\small,
left=6pt, right=6pt, top=6pt, bottom=6pt
]
You are an expert in creating combinatorics math problems. Your task is to rewrite a given math problem to make it more engaging, story-like, or similar in style to the provided reference problems, while STRICTLY preserving the underlying mathematical logic and constraints.

\vspace{0.8em}
\textbf{Here are some reference problems that have a similar mathematical structure or style. Use them as inspiration for the tone and narrative style:}
---
\textcolor{blue}{\{references\_text\}}
---

\textbf{Here is the mathematical logic of the original problem (represented in Cofola language):}
\begin{tcolorbox}[colback=white, boxrule=0.5pt, sharp corners, left=2pt, top=2pt, bottom=2pt]
\textcolor{blue}{\{original\_cofola\}}
\end{tcolorbox}
(Note: The rewritten problem MUST correspond exactly to this logic. Do not change numbers, constraints, or the type of combinatorial object being counted.)

\vspace{0.8em}
\textbf{Original Problem Text:}
``\textcolor{blue}{\{original\_question\}}''

\vspace{0.8em}
\textbf{Task:}
Rewrite the ``Original Problem Text'' above.
\begin{enumerate}[leftmargin=*, nosep]
\item Keep the mathematical core exactly the same (same numbers, same conditions).
\item You can change the setting (e.g., from balls in boxes to students in classrooms) if it fits the logic, or just improve the narrative of the current setting.
\item Ensure the question is clear and unambiguous.
\end{enumerate}

\textbf{Output Format:}
Just provide the rewritten problem text. Do not include explanations or the Cofola code.

\textbf{Rewritten Problem:}
\end{tcolorbox}
\caption{The prompt used to generate rewritten variations of combinatorial problems. Placeholders in blue are replaced with reference examples, the Cofola logic representation, and the original problem text for each instance.}
\label{fig:rewrite_prompt}
\end{figure*}

\begin{figure*}[ht]
\centering
\begin{tcolorbox}[
colback=gray!5,
colframe=black!70,
boxrule=0.8pt,
arc=2mm,
title=\textbf{Prompt Template for Natural Language to Cofola Translation},
fonttitle=\bfseries\sffamily,
fontupper=\ttfamily\small,
left=6pt, right=6pt, top=6pt, bottom=6pt
]
\textbf{\# Role}
You are an expert combinatorial problem solver and an assistant strictly trained to translate natural language combinatorial questions into a specialized domain-specific language (DSL) called \textbf{Cofola}.

\vspace{0.5em}
\textbf{\# Task}
Your task is to read a combinatorics problem described in English and map it line-by-line or step-by-step into Cofola code. You must use ONLY the mappings provided in the dictionary below.

\vspace{0.8em}
\textbf{\# Dictionary: Cofola to Natural Language Mapping}

\textit{[The full dictionary defines exhaustive one-to-one mappings between natural language patterns and Cofola constructs across the following categories: (1) Types \& Declarations; (2) Operations (Set \& Bag Math, Choice \& Arrangement, Partitions \& Groupings); (3) Constraints (comparators, general constraint mappings, and sequence/tuple-specific constraints); (4) Question Template (target output format). See the supplementary material for the complete dictionary.]}

\vspace{0.8em}
\textbf{Here are some reference examples showing the expected input-output format:}
---
\textcolor{blue}{\{references\_text\}}
---

\vspace{0.5em}
\textbf{Important requirements:}
\begin{itemize}[leftmargin=*, nosep]
\item Analyze the reference examples carefully to understand the expected code structure and style
\item Generate code that exactly matches the format shown in the examples
\item Output only the program code (Cofola DSL) without any explanations, comments or additional markdown text like \texttt{```cofola}
\item Ensure the code follows proper syntax and can be executed correctly
\end{itemize}

\vspace{0.8em}
\textbf{Task:}
Translate the following natural language problem into a Cofola program.

\vspace{0.3em}
\textbf{Problem:}
\textcolor{blue}{\{original\_question\}}

\vspace{0.3em}
\textbf{Program:}
\end{tcolorbox}
\caption{The prompt template used for translating natural language combinatorial problems into Cofola DSL code. The Dictionary section (summarized above) provides exhaustive one-to-one mappings between natural language patterns and Cofola constructs; the full dictionary is available in the supplementary material. Placeholders in blue are replaced with retrieved reference examples and the target problem.}
\label{fig:nl_to_cofola_prompt}
\end{figure*}

\begin{figure*}[ht]
  \centering
  \begin{minipage}[t]{0.48\textwidth}
    \begin{tcolorbox}[
      colback=blue!5,
      colframe=blue!70,
      boxrule=0.8pt,
      arc=2mm,
      title=\textbf{Example 1: Choose (Level 2)},
      fonttitle=\bfseries\sffamily,
      width=\textwidth,
      height=5.5cm,
    ]
    \textbf{Problem:} My school's math club has 6 boys and 8 girls. 
    I need to select a team to send to the state math competition. 
    We want 6 people on the team. In how many ways can I select the 
    team without restrictions?

    \vspace{0.5em}
    \textbf{Cofola Encoding:}
    \begin{verbatim}
    boys = set(boy0...6)
    girls = set(girl0...8)
    team = choose(boys + girls, 6)
    \end{verbatim}
    \end{tcolorbox}
  \end{minipage}
  \hfill
  \begin{minipage}[t]{0.48\textwidth}
    \begin{tcolorbox}[
      colback=blue!5,
      colframe=blue!70,
      boxrule=0.8pt,
      arc=2mm,
      title=\textbf{Example 2: Group (Level 5)},
      fonttitle=\bfseries\sffamily,
      width=\textwidth,
      height=5.5cm,
    ]
    \textbf{Problem:} How many ways are there to put 4 balls in 3 boxes 
    if two balls are indistinguishably green, two are indistinguishably 
    red, and the boxes are distinguishable?

    \vspace{0.5em}
    \textbf{Cofola Encoding:}
    \begin{verbatim}
    balls = bag(gree: 2, red: 2)
    boxes = Group(balls, 3)
    \end{verbatim}
    \end{tcolorbox}
  \end{minipage}

  \caption{Example problems from the MATH dataset with their formal 
           Cofola encodings used as references during the rewriting 
           process, where \texttt{bag} is the Cofola construct for 
           multisets.}
  \label{fig:math_examples}
\end{figure*}

Figure~\ref{fig:rewrite_prompt} presents the prompt template used for problem rewriting. 
The prompt instructs the model to generate diverse, story-like reformulations while strictly preserving the underlying combinatorial logic specified in the Cofola representation. 
By explicitly separating mathematical constraints from surface-level language and providing reference problems as stylistic guidance, 
this design enables controlled variation in problem phrasing without altering the solution space.

Figure~\ref{fig:nl_to_cofola_prompt} presents the prompt template used for natural-language-to-Cofola back-translation. 
This prompt asks the model to recover the formal Cofola program from a rewritten natural-language problem using an explicit dictionary of mappings between linguistic patterns and Cofola constructs. 
The generated Cofola program is then executed by the solver and compared against the original ground-truth answer. 
Together, the rewriting prompt and the back-translation prompt form a closed verification loop, allowing us to study reasoning robustness under paraphrasing while avoiding confounding changes to problem structure.


\begin{figure*}[ht]
  \centering
  \begin{minipage}[t]{0.48\textwidth}
    \begin{tcolorbox}[
      colback=blue!5,
      colframe=blue!70,
      boxrule=0.8pt,
      arc=2mm,
      title=\textbf{Template-based},
      fonttitle=\bfseries\sffamily,
      width=\textwidth,
      height=4.9cm, 
    ]
    Set $\mathit{set}_0$ contains: $\mathit{e}_8$, $\mathit{e}_6$, 
    $\mathit{e}_2$, $\mathit{e}_9$, $\mathit{e}_7$, 
    $\mathit{e}_4$, $\mathit{element}_5$, $\mathit{e}_1$, 
    $\mathit{e}_3$, $\mathit{e}_{10}$.
    
    From $\mathit{set}_0$, we select 7 elements without replacement 
    to form $\mathit{choose}_0$.
    
    How many different ways can we obtain $\mathit{choose}_0$? 

    Count each distinct combination of $\mathit{choose}_0$ as one way.
    \end{tcolorbox}
  \end{minipage}
  \hfill
  \begin{minipage}[t]{0.48\textwidth}
    \begin{tcolorbox}[
      colback=blue!5,
      colframe=blue!70,
      boxrule=0.8pt,
      arc=2mm,
      title=\textbf{Rewritten\textsuperscript{*}},
      fonttitle=\bfseries\sffamily,
      width=\textwidth,
      height=4.2cm, 
    ]
    A school's debate team consists of 10 students. 
    For an upcoming national competition, the coach needs 
    to select a group of 7 students to represent the school.

    In how many different ways can the coach choose the 
    7-member team from the 10 available students?
    \end{tcolorbox}
  \end{minipage}

  \vspace{4pt}
  \raggedright
  \textsuperscript{*}Rewritten by \texttt{gemini-3-flash-preview-thinking}.
  \caption{An example of a generated combinatorial problem in its 
           template-based and rewritten natural-language forms. 
           Both formulations encode the same underlying 
           combinatorial structure (Choose from a set of 10 
           elements without replacement, selecting 7).}
  \label{fig:problem-examples}
\end{figure*}

\section{Template}
\label{app:Template_section}

We list the template system we design to verbalize generated Cofola instances as natural-language problems. 
The template system separates problem statements into four components: entity/object declarations, set and bag construction steps, typed combinatorial-object operations, and constraints.
This separation allows the generator to preserve the formal structure of each instance while producing readable natural-language descriptions.

Templates are parameterized by placeholders enclosed in braces, such as \{set\_name\}, \{source\_name\}, \{node\_name\}, and \{k\}. 
During generation, these placeholders are filled with sampled entities, intermediate variable names, operation parameters, and constraint values. 
For example, the template ``From \{source\_name\}, we select \{k\} elements without replacement to form \{node\_name\}.'' can be instantiated as ``From aux\_0, we select 3 elements without replacement to form aux\_1.'' 
The same mechanism is used for set/multiset construction, sequence constraints, tuple constraints, and final counting questions.

Table~\ref{tab:entity-universe} gives templates for entity declarations and set/bag construction; Table~\ref{tab:combinatorial-ops} lists typed object-operation templates; Tables~\ref{tab:constraints-general-Group-bag}, \ref{tab:constraints-sequence}, and~\ref{tab:constraints-tuple} summarize constraint templates; and Table~\ref{tab:question-comparator} reports the question, comparator, polarity, and entity-pool templates.

\begin{table*}[ht]
\centering
\caption{Math templates for entity type declarations and universe construction.}
\label{tab:entity-universe}
\small
\begin{tabularx}{\textwidth}{lX}
\toprule
\textbf{Category} & \textbf{Template} \\
\midrule
\multicolumn{2}{l}{\textbf{Entity Type Declarations}} \\
\midrule
Set & \texttt{Set \{set\_name\} contains: \{description\}.} \\
Bag (Multiset) & \texttt{\{set\_name\} contains: \{description\}.} \\
Tuple & \texttt{\{set\_name\} contains: \{description\}.} \\
Sequence & \texttt{\{set\_name\} contains: \{description\}.} \\
\midrule
\multicolumn{2}{l}{\textbf{Universe Construction Operations}} \\
\midrule
\texttt{SET\_UNION} & We take the union of \texttt{\{left\_name\}} and \texttt{\{right\_name\}}, calling the result \texttt{\{node\_name\}}. \\
\texttt{SET\_INTERSECTION} & We take the intersection of \texttt{\{left\_name\}} and \texttt{\{right\_name\}}, calling the result \texttt{\{node\_name\}}. \\
\texttt{SET\_DIFFERENCE} & Starting from \texttt{\{left\_name\}}, we remove elements of \texttt{\{right\_name\}} to get \texttt{\{node\_name\}}. \\
\texttt{BAG\_UNION} & We merge \texttt{\{left\_name\}} and \texttt{\{right\_name\}} by adding their quantities to obtain \texttt{\{node\_name\}}. \\
\texttt{BAG\_INTERSECTION} & We combine \texttt{\{left\_name\}} and \texttt{\{right\_name\}} by taking the minimum quantity of each element to get \texttt{\{node\_name\}}. \\
\texttt{BAG\_DIFFERENCE} & Starting from \texttt{\{left\_name\}}, we subtract the quantities in \texttt{\{right\_name\}} to get \texttt{\{node\_name\}}, ensuring no element's quantity becomes negative. \\
\bottomrule
\end{tabularx}
\end{table*}

\begin{table*}[ht]
\centering
\caption{Math templates for combinatorial operations.}
\label{tab:combinatorial-ops}
\small
\begin{tabularx}{\textwidth}{lX}
\toprule
\textbf{Operation} & \textbf{Template} \\
\midrule
\multicolumn{2}{l}{\textbf{CHOOSE}} \\
\midrule
\texttt{CHOOSE} & From \texttt{\{source\_name\}}, we select \texttt{\{k\}} elements without replacement to form \texttt{\{node\_name\}}. \\
\texttt{CHOOSE\_REPLACE} & From \texttt{\{source\_name\}}, we repeatedly select \texttt{\{k\}} elements (with replacement) to form \texttt{\{node\_name\}}. \\
\midrule
\multicolumn{2}{l}{\textbf{Group}} \\
\midrule
\texttt{COMPOSE} & We divide \texttt{\{source\_name\}} into \texttt{\{k\}} groups (allowing empty groups and groups are labeled), storing the result in \texttt{\{node\_name\}}. \\
\texttt{PARTITION} & We split \texttt{\{source\_name\}} into \texttt{\{k\}} groups (allowing empty groups and groups are not labeled), storing the result in \texttt{\{node\_name\}}. \\
\midrule
\multicolumn{2}{l}{\textbf{TUPLE}} \\
\midrule
\texttt{CHOOSE\_REPLACE\_TUPLE} & We draw \texttt{\{k\}} elements from \texttt{\{source\_name\}} (with replacement) and arrange them in \texttt{\{node\_name\}}. \\
\texttt{CHOOSE\_TUPLE} & From \texttt{\{source\_name\}}, we pick \texttt{\{k\}} elements and arrange them in order to form \texttt{\{node\_name\}}. \\
\texttt{TUPLE} & We convert \texttt{\{source\_name\}} into an ordered sequence \texttt{\{node\_name\}}. \\
\midrule
\multicolumn{2}{l}{\textbf{SEQUENCE}} \\
\midrule
\texttt{SEQUENCE} & We arrange all elements from \texttt{\{source\_name\}} to form an ordered sequence \texttt{\{node\_name\}}. \\
\texttt{CHOOSE\_REPLACE\_SEQUENCE} & We draw \texttt{\{k\}} elements from \texttt{\{source\_name\}} with replacement and arrange them in sequence \texttt{\{node\_name\}}. \\
\texttt{CIRCLE} & We arrange all elements from \texttt{\{source\_name\}} around a circle called \texttt{\{node\_name\}}\texttt{\{reflection\}}. \\

\bottomrule
\end{tabularx}
\end{table*}

\begin{table*}[ht]
\centering
\caption{Math templates for constraints (Part 1): General and Group constraints.}
\label{tab:constraints-general-Group-bag}
\small
\begin{tabularx}{\textwidth}{lX}
\toprule
\textbf{Constraint Name} & \textbf{Template} \\
\midrule
\multicolumn{2}{l}{\textbf{General Constraints}} \\
\midrule
\texttt{MemberConstraint} & \texttt{\{target\} \{positive\}contains \{entity\}.} \\
\texttt{DisjointConstraint} & \texttt{\{second\_set\} is disjoint from (\{first\_set\}).} \\
\texttt{SubsetConstraint} & \texttt{\{first\_set\} is a subset of (\{second\_set\}).} \\
\texttt{EqualityConstraint} & \texttt{\{left\_name\} must be the same as \{right\_name\}.} \\
\texttt{CardinalityConstraint} & \texttt{\{target\} must contain \{comparator\} \{value\} elements.} \\
\texttt{CountConstraint} & The element \texttt{\{entity\}} must appear \texttt{\{comparator\} \{value\}} times in \texttt{\{target\}}. \\
\texttt{NonEmptyConstraint} & \texttt{\{target\} must not be empty.} \\
\texttt{LinearCardinalityConstraint} & The total size of \texttt{\{left\}} and \texttt{\{right\}} must be \texttt{\{comparator\} \{value\}}. \\
\texttt{IndexMemberConstraint} & The set at index \texttt{\{index\}} of \texttt{\{target\} \{positive\}} belong to \texttt{\{container\}}. \\
\texttt{IndexEqualMemberConstraint} & The element at index \texttt{\{index\}} of \texttt{\{target\} \{positive\}} be \texttt{\{entity\}}. \\
\midrule
\multicolumn{2}{l}{\textbf{Group Constraints}} \\
\midrule
\texttt{ComposeSizeConstraint} & The size of the \texttt{\{index\}}th subset must be \texttt{\{comparator\} \{param\}}. \\
\texttt{QuantifiedConstraint} & For each group in the \texttt{\{target\}}, the number of entities it contains must \texttt{\{comparator\} \{value\}}. \\

\bottomrule
\end{tabularx}
\end{table*}

\begin{table*}[ht]
\centering
\caption{Math templates for constraints (Part 2): Sequence ordering constraints.}
\label{tab:constraints-sequence}
\small
\begin{tabularx}{\textwidth}{lX}
\toprule
\textbf{Constraint Name} & \textbf{Template} \\
\midrule
\multicolumn{2}{l}{\textbf{Sequence Ordering Constraints (Standard)}} \\
\midrule
\texttt{PredecessorConstraint} (pos) & Sequence \texttt{\{target\}} must contain \texttt{\{entity1\}} and \texttt{\{entity2\}} such that \texttt{\{entity1\}} is immediately before \texttt{\{entity2\}}. \\
\texttt{PredecessorConstraint} (neg) & Sequence \texttt{\{target\}} does not have \texttt{\{entity1\}} immediately before \texttt{\{entity2\}} if both are present. \\
\texttt{LessThanConstraint} (pos) & Sequence \texttt{\{target\}} must contain \texttt{\{entity1\}} and \texttt{\{entity2\}} such that \texttt{\{entity1\}} comes before \texttt{\{entity2\}} (not necessarily immediately). \\
\texttt{LessThanConstraint} (neg) & Sequence \texttt{\{target\}} does not have \texttt{\{entity1\}} before \texttt{\{entity2\}} (immediately or not) if both are present. \\
\texttt{NextToConstraint} (pos) & Sequence \texttt{\{target\}} must contain \texttt{\{entity1\}} and \texttt{\{entity2\}} such that they are adjacent. \\
\texttt{NextToConstraint} (neg) & Sequence \texttt{\{target\}} does not have \texttt{\{entity1\}} and \texttt{\{entity2\}} adjacent if both are present. \\
\texttt{TogetherConstraint} (pos) & Entities (\texttt{\{group\}}) must be together in sequence \texttt{\{target\}} if any are present. \\
\texttt{TogetherConstraint} (neg) & Sequence \texttt{\{target\}} contains the entities (\texttt{\{group\}}) and they are not together. \\
\midrule
\multicolumn{2}{l}{\textbf{Sequence Ordering Constraints (Bag Semantics)}} \\
\midrule
\texttt{PredecessorBagConstraint} (pos) & There is a pair of \texttt{\{entity1\}} and \texttt{\{entity2\}} in sequence \texttt{\{target\}} such that \texttt{\{entity1\}} is immediately before \texttt{\{entity2\}}. \\
\texttt{PredecessorBagConstraint} (neg) & Sequence \texttt{\{target\}} does not have any \texttt{\{entity1\}} immediately before any \texttt{\{entity2\}} (\texttt{\{target\}} that contain only \texttt{\{entity1\}} or only \texttt{\{entity2\}} or none of them are also allowed). \\
\texttt{LessThanBagConstraint} (pos) & There is a pair of \texttt{\{entity1\}} and \texttt{\{entity2\}} in sequence \texttt{\{target\}} such that \texttt{\{entity1\}} comes before \texttt{\{entity2\}} (not necessarily immediately). \\
\texttt{LessThanBagConstraint} (neg) & Sequence \texttt{\{target\}} does not have any \texttt{\{entity1\}} before any \texttt{\{entity2\}} (immediately or not) (\texttt{\{target\}} that contain only \texttt{\{entity1\}} or only \texttt{\{entity2\}} or none of them are also allowed). \\
\texttt{NextToBagConstraint} (pos) & There is a pair of \texttt{\{entity1\}} and \texttt{\{entity2\}} in sequence \texttt{\{target\}} such that they are adjacent. \\
\texttt{NextToBagConstraint} (neg) & Sequence \texttt{\{target\}} does not have any \texttt{\{entity1\}} and \texttt{\{entity2\}} adjacent (\texttt{\{target\}} that contain only \texttt{\{entity1\}} or only \texttt{\{entity2\}} or none of them are also allowed). \\
\midrule
\multicolumn{2}{l}{\textbf{Complex Sequence Count Constraints}} \\
\midrule
\texttt{SequenceCountConstraint} & In sequence \texttt{\{target\}}, \texttt{\{count\_type\}}. \\
\quad \textit{next\_to} & \texttt{\{entity1\}} and \texttt{\{entity2\}} are adjacent (in any order) \texttt{\{comparator\} \{value\}} times \\
\quad \textit{predecessor} & \texttt{\{entity1\}} is immediately followed by \texttt{\{entity2\} \{comparator\} \{value\}} times \\
\quad \textit{less\_than} & \texttt{\{entity1\}} is followed by \texttt{\{entity2\}} (allowing gaps) \texttt{\{comparator\} \{value\}} times \\
\bottomrule
\end{tabularx}
\end{table*}

\begin{table*}[ht]
\centering
\caption{Math templates for constraints (Part 3): Tuple-specific constraints.}
\label{tab:constraints-tuple}
\small
\begin{tabularx}{\textwidth}{lX}
\toprule
\textbf{Constraint Name} & \textbf{Template} \\
\midrule
\texttt{TupleMembershipConstraint} & \texttt{\{target\} \{positive\} contains \{entity\}.} \\
\texttt{TupleIndexMembershipConstraint} & The element at position \texttt{\{index\}} of \texttt{\{target\}} must \texttt{\{positive\}be \{entity\}}. \\
\texttt{TupleAllMembersInConstraint} & All elements in \texttt{\{target\}} must belong to \texttt{\{target\_set\}}. \\
\texttt{TupleCountSizeConstraint} & The number of \texttt{\{count\_obj\}} in \texttt{\{target\}} must be \texttt{\{comparator\} \{param\}}. \\
\texttt{TupleIndexEqConstraint} & The \texttt{\{index\}}th element of \texttt{\{target\} \{positive\}is \{entity\}}. \\
\texttt{TupleDedupCountSizeConstraint} & In \texttt{\{target\}}, the number of distinct elements belonging to \texttt{\{count\_obj\}} must be \texttt{\{comparator\} \{param\}}. \\
\texttt{PositionConstraint} & The \texttt{\{ordinal\}} position in \texttt{\{target\}} must be \texttt{\{entity\}}. \\
\bottomrule
\end{tabularx}
\end{table*}

\begin{table*}[ht]
\centering
\caption{Question templates and comparator mappings for natural language generation.}
\label{tab:question-comparator}
\small
\begin{tabularx}{\textwidth}{lX}
\toprule
\textbf{Category} & \textbf{Template / Mapping} \\
\midrule
\multicolumn{2}{l}{\textbf{Default Question Template}} \\
\midrule
Counting question & How many different ways can we obtain \texttt{\{uncertain\_node\_name\}}? Count each distinct combination of \texttt{\{uncertain\_node\_name\}} as one way. \\
\midrule
\multicolumn{2}{l}{\textbf{Comparator to Natural Language (\texttt{comparator2str})}} \\
\midrule
\texttt{==} & exactly \\
\texttt{<=} & at most \\
\texttt{<}  & less than \\
\texttt{>=} & at least \\
\texttt{>}  & greater than \\
\texttt{!=} & not exactly \\
\midrule
\multicolumn{2}{l}{\textbf{Positive/Negative Boolean Mapping (\texttt{positive\_bool2txt})}} \\
\midrule
\texttt{true}  & (empty string) \\
\texttt{false} & \texttt{not} \\
\midrule
\multicolumn{2}{l}{\textbf{Special Tokens}} \\
\midrule
\texttt{entity\_type} & \texttt{element} \\
\texttt{entity\_unit} & (single space) \\
\texttt{empty\_entity} & \texttt{no \{entity\_type\} at all} \\
\bottomrule
\end{tabularx}
\end{table*}

\section{Error Examples of \textbf{gpt-5.5}}
\label{app:error_analysis_section}

We present representative error examples generated by \textbf{gpt-5.5} in \Cref{err:together_block_perm,err:negated_together,err:independent_sample,err:circle_multiset,err:chained_together,err:compound_block}.
The text in red indicates the parts where the model made a mistake.
We note that we do not cover all error types here as exhaustive manual inspection is labor-intensive.
Now, let us analyze these error examples in detail.
\begin{itemize}
  \item \Cref{err:together_block_perm} shows a circle arrangement where the model treats a ``together'' block as a fixed ordered chunk, omitting the $k!$ internal permutations of the block and undercounting by a factor of $3!$.
  \item \Cref{err:negated_together} shows a sequence problem where the model reads ``\texttt{together} not in seq'' as ``no two elements of the set are adjacent'', a much stricter condition than the intended ``the elements do not all form one contiguous block''.
  \item \Cref{err:independent_sample} illustrates a case where two independent \texttt{choose\_tuple} operators on the same bag are treated by the model as a single joint deal that depletes the bag, rather than two independent draws from the original bag.
  \item \Cref{err:circle_multiset} shows a multiset circular arrangement where the model uses the linear formula $n!/\prod m_i!$ instead of the circular formula $(n{-}1)!/\prod m_i!$, overcounting by a factor of $n$.
  \item \Cref{err:chained_together} shows a circle problem with two \texttt{together} constraints sharing an entity. The model correctly identifies the forced chain but treats it as a single fixed orientation, missing the four internal orientations that the chain admits.
  \item \Cref{err:compound_block} shows a circle problem combining \texttt{together} (over three entities) with \texttt{next\_to}. The combined block has four distinct configurations (two internal orderings $\times$ two chain directions in a rotation-only circle), all of which the model collapses to one.
\end{itemize}

The first two patterns reflect a systematic gap between the natural-language reading of Cofola constraints and their formal semantics over multisets, particularly the symmetry of ``together'' blocks. The remaining patterns reflect classical combinatorial slips that scale with the structure of compound constraints.
We anticipate that explicitly formalizing these semantic edge cases in the prompt template, or pairing the LLM with a symbolic verifier, would mitigate a substantial fraction of these errors~\citep{gao2023pal}.

\begin{figure*}[t]
\centering
\begin{errexample}{Missing internal permutations of a ``together'' block}{together_block_perm}
\bgroup\obeylines
\textbf{Question:}

Set $set_0$ contains: element7, element5, element8, element6, element2, element4, element1, element3.
We arrange all elements from $set_0$ around a circle called $circle_0$.
Entities (element2, element3, element5) must be together in sequence $circle_0$ if any are present.
How many different ways can we obtain $circle_0$? Count each distinct combination of $circle_0$ as one way.

\textbf{Reference: `720`}

\textbf{Prediction:}

\textcolor{red}{Treat (element2, element3, element5) as one fixed consecutive block}, since they must be together in that sequence.

So we arrange the following 6 objects around a circle:

$[\text{element2}, \text{element3}, \text{element5}],\ \text{element1},\ \text{element4},\ \text{element6},\ \text{element7},\ \text{element8}.$

The number of circular arrangements of 6 distinct objects is $(6-1)! = 5! = 120$.

\boxed{120}
\egroup
\end{errexample}
\end{figure*}

\begin{figure*}[t]
\centering
\begin{errexample}{Misreading negated ``together'' as ``no two adjacent''}{negated_together}
\bgroup\obeylines
\textbf{Question:}

$bag_0$ contains: element3:2, element5:2, element6:2.
We arrange all elements from $bag_0$ to form an ordered sequence $sequence_0$.
Sequence $sequence_0$ contains the entities (element3, element5) and they are not together.
How many different ways can we obtain $sequence_0$? Count each distinct combination of $sequence_0$ as one way.

\textbf{Reference: `72`}

\textbf{Prediction:}

Let $A=\text{element3}$, $B=\text{element5}$, $C=\text{element6}$. We arrange the multiset $\{A,A,B,B,C,C\}$ with the condition that $A$ and $B$ are not together, \textcolor{red}{meaning no $A$ is adjacent to any $B$}.

Use the two $C$'s as separators: $\_\ C\ \_\ C\ \_$.
There are 3 gaps where the $A$'s and $B$'s may be placed.
To avoid adjacency between $A$ and $B$, each gap can contain only $A$'s, only $B$'s, or be empty.
We distribute 2 copies of $A$ and 2 copies of $B$ among these 3 gaps:
\begin{itemize}
  \item $A$'s in 1 gap and $B$'s in 1 gap: $3\times 2 = 6$.
  \item $A$'s in 1 gap and $B$'s split across 2 gaps: $3$.
  \item $A$'s split across 2 gaps and $B$'s in 1 gap: $3$.
\end{itemize}
Total: $6+3+3 = 12$.

\boxed{12}
\egroup
\end{errexample}
\end{figure*}









\begin{figure*}[t]
\centering
\begin{errexample}{Joint vs.\ independent sampling from the same bag}{independent_sample}
\bgroup\obeylines
\textbf{Question:}

$bag_0$ contains: element2:2, element5:1, element6:3.
From $bag_0$, we pick 2 elements and arrange them in order to form $choose\_tuple_0$.
From $bag_0$, we pick 3 elements and arrange them in order to form $choose\_tuple_1$.
How many different ways can we obtain $choose\_tuple_0, choose\_tuple_1$? Count each distinct combination as one way.

\textbf{Reference: `152`}

\textbf{Prediction:}

The bag contains a multiset of 6 elements: $\{2,2,5,6,6,6\}$.
\textcolor{red}{Together, the two tuples use $2+3=5$ elements from the bag, so this is equivalent to forming an ordered sequence of length 5 from the bag (first 2 positions = $choose\_tuple_0$, last 3 = $choose\_tuple_1$).}

Since we use 5 out of 6 elements, exactly one element is left unused.
\begin{itemize}
  \item Leave out one 2: counts $(1,1,3) \Rightarrow 5!/(1!\,1!\,3!) = 20$.
  \item Leave out the 5: counts $(2,0,3) \Rightarrow 5!/(2!\,3!) = 10$.
  \item Leave out one 6: counts $(2,1,2) \Rightarrow 5!/(2!\,1!\,2!) = 30$.
\end{itemize}
Total: $20 + 10 + 30 = 60$.

\boxed{60}
\egroup
\end{errexample}
\end{figure*}

\begin{figure*}[t]
\centering
\begin{errexample}{Missing rotational quotient for a multiset circle}{circle_multiset}
\bgroup\obeylines
\textbf{Question:}

$bag_0$ contains: element1:1, element2:1, element5:3, element6:1, element7:1, element9:1, element10:2.
We arrange all elements from $bag_0$ around a circle called $circle_0$.
How many different ways can we obtain $circle_0$? Count each distinct combination of $circle_0$ as one way.

\textbf{Reference: `30240`}

\textbf{Prediction:}

The bag has 10 elements with multiplicities $\{1,1,3,1,1,1,2\}$.
\textcolor{red}{The number of distinguishable circular arrangements of the multiset equals the multinomial coefficient}
$\frac{10!}{1!\,1!\,3!\,1!\,1!\,1!\,2!} = \frac{3{,}628{,}800}{12} = 302{,}400.$

\boxed{302400}
\egroup
\end{errexample}
\end{figure*}

\begin{figure*}[t]
\centering
\begin{errexample}{Chained ``together'' constraints sharing an entity}{chained_together}
\bgroup\obeylines
\textbf{Question:}

Set $set_0$ contains: element7, element5, element8, element6, element2, element4, element1, element3.
We arrange all elements from $set_0$ around a circle called $circle_0$, two arrangements are considered the same if one can be rotated or reflected (flipped) to match the other.
Entities (element6, element7) must be together in sequence $circle_0$ if any are present.
Entities (element5, element2, element7) must be together in sequence $circle_0$ if any are present.
How many different ways can we obtain $circle_0$? Count each distinct combination of $circle_0$ as one way.

\textbf{Reference: `48`}

\textbf{Prediction:}

The required ``together'' of (element5, element2, element7) means these three elements must be consecutive around the circle. Also, (element6, element7) must be consecutive, so element7 must be adjacent to both element2 and element6.

\textcolor{red}{So the forced block is the chain element5--element2--element7--element6 up to reflection.}

\textcolor{red}{Treating this chain as one single block, we need to arrange the following 5 objects around a circle: $[\text{element5}, \text{element2}, \text{element7}, \text{element6}]$, element1, element3, element4, element8. The number of circular arrangements of 5 distinct objects up to rotation and reflection is $(5{-}1)!/2 = 12$.}

\boxed{12}
\egroup
\end{errexample}
\end{figure*}

\begin{figure*}[t]
\centering
\begin{errexample}{Compound block from \texttt{together} and \texttt{next\_to} in a circle}{compound_block}
\bgroup\obeylines
\textbf{Question:}

Set $set_0$ contains: element7, element5, element8, element6, element11, element12, element9, element2, element4, element1, element3, element10.
We arrange all elements from $set_0$ around a circle called $circle_0$.
Entities (element12, element6, element9) must be together in $circle_0$ if any are present.
$circle_0$ must contain element12 and element11 such that they are adjacent.
$circle_0$ does not have element2 and element3 adjacent if both are present.
How many different ways can we obtain $circle_0$?

\textbf{Reference: `120960`}

\textbf{Prediction:}

We treat (element12, element6, element9) as a required consecutive sequence. Since element12 must also be adjacent to element11, and element12 is already adjacent to element6 in the required sequence, element11 must be on the other side of element12. \textcolor{red}{Thus we form one block (element11, element12, element6, element9).}

Now we arrange this block together with the remaining 8 elements, giving $9$ objects around a circle: $(9-1)! = 8! = 40320$. Subtract arrangements where element2 and element3 are adjacent (treat them as one block of size 2): $2! \cdot 7! = 10080$.

\textcolor{red}{Valid arrangements: $40320 - 10080 = 30240$.}

\boxed{30240}
\egroup
\end{errexample}
\end{figure*}

\end{document}